\documentclass[preprint,12pt]{elsarticle}

\usepackage{amssymb}
\usepackage{amsmath}
\usepackage{comment}
\usepackage{booktabs} 
\usepackage{multirow}
\usepackage{graphicx}
\usepackage{textcomp}
\usepackage{multirow}  
\usepackage{hyperref}
\usepackage{color}
\definecolor{Blue}{rgb}{0.3,0.3,0.9}
\usepackage{xcolor}
\usepackage{subcaption}
\usepackage{pgfplots}
\pgfplotsset{compat=1.18}
\usepackage{tikz}
\usepackage{caption}

\journal{Medical Image Analysis}

\begin{document}

\begin{frontmatter}



\title{Example-based Robust Abnormality Detection with Minimal Annotations using Exemplar Med-DETR}

\author[label1,label2]{Sheethal Bhat}\ead{sheethal.bhat@fau.de}
\author[label3]{Bogdan Georgescu}
\author[label3]{Awais Mansoor}
\author[label1]{Mathias Zinnen}
\author[label3]{Pranjal Sahu}
\author[label2]{Florin C. Ghesu}
\author[label3]{Sasa Grbic}
\author[label1]{Andreas Maier}
\affiliation[label1]{organization={Pattern Recognition Lab, Friedrich-Alexander-Universität},
            city={Erlangen},
            postcode={91058}, 
            country={Germany}}
\affiliation[label2]{organization={Digital Technology and Innovation, Siemens Healthineers},
            city={Erlangen},
            postcode={},
            country={Germany}}
\affiliation[label3]{organization={Digital Technology and Innovation, Siemens Medical Solutions},
            city={Princeton}, 
            state={NJ},
            postcode={08540},
            country={U.S.A}}
            
\begin{abstract}
Reducing annotation requirements remains a key challenge in developing robust medical object detectors. To address this, Vision-Language (VL) object detection methods leverage grounding text information to enable powerful zero-shot and few-shot object detectors in the natural image domain \cite{liu2024grounding,li2022grounded,zareian2021open,minderer2022simple}. However, transferring these methods to the medical domain is challenging due to the absence of comparable quality and quantity of the grounding data. Regardless, significant contextual and non-imaging information exists in medical images that remains underutilized. Few-shot learning (FSL) techniques partially address this limitation but struggle to generalize to unseen medical findings and require extensive retraining when new findings are introduced \cite{bulat2023fs,zhang2022meta}.
To overcome these challenges, we extend our prior EM-DETR framework \cite{emdetr} and introduce a scalable FS detection approach designed for efficient abnormality detection in Chest X-Ray (CXR) images under minimal supervision. The proposed architecture incorporates exemplar-based feature generation and domain-aware contrastive optimization, enabling effective adaptation to novel disease findings without exhaustive retraining. Our method achieves near state-of-the-art (SOTA) detection performance using less than 10\% of the annotated data, demonstrating its potential for practical, annotation-efficient clinical deployment across both proprietary and public CXR datasets.

\end{abstract}

\begin{graphicalabstract}
\begin{figure}[ht]
    \centering
    \includegraphics[width=0.7\linewidth]{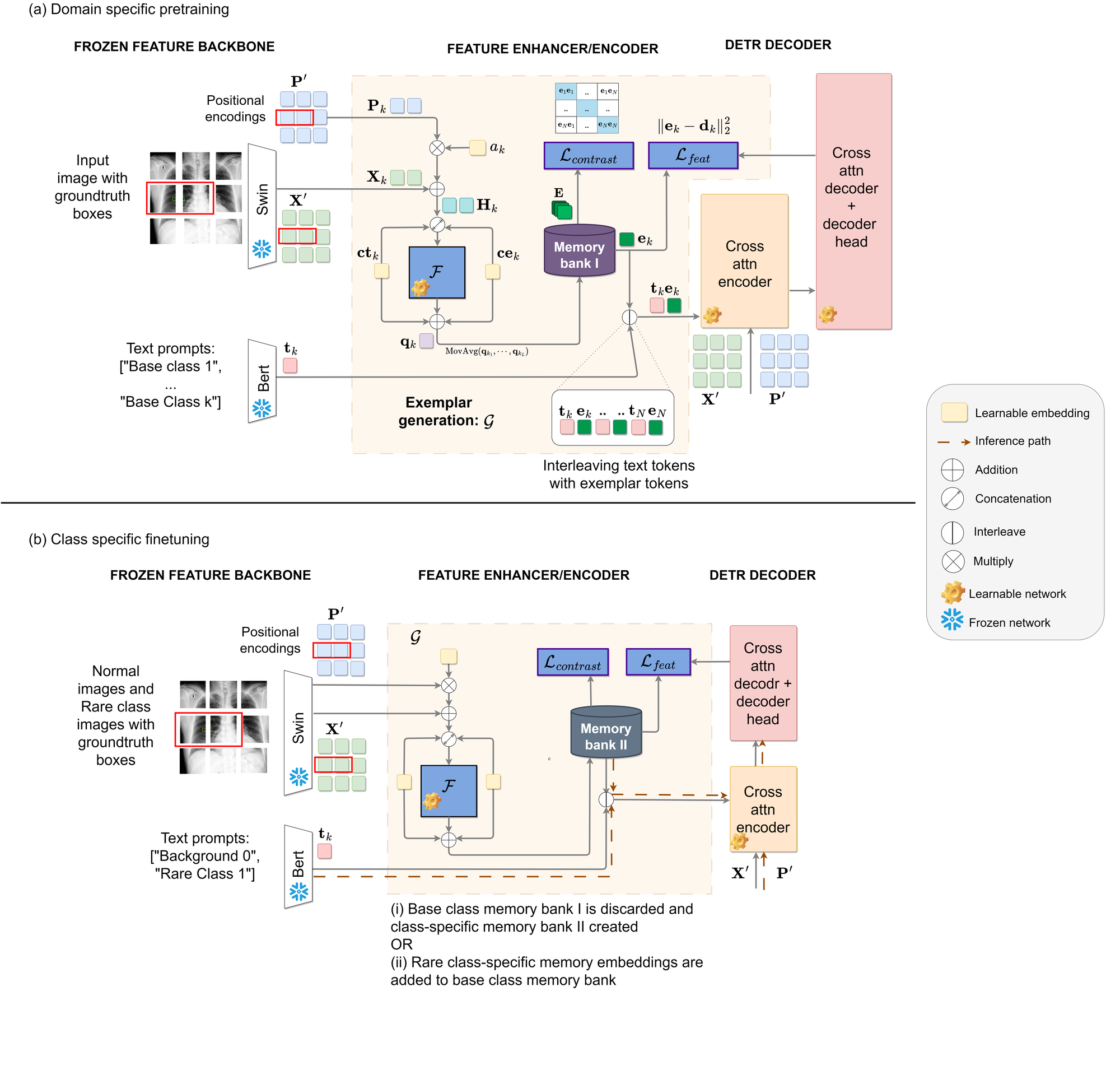}
    \caption{Overview of the pretraining and finetuning stages of Exemplar Med-DETR (EM-DETR). The model is pretrained on a set of base classes and finetuned on a smaller set of the novel class of interest. The Exemplar generation module $\mathcal{G}$ extracts the patches belonging to Region-of-Interest (ROI) and computes a unique exemplar or prototype embedding per class. These embeddings are stored in a memory bank and are used during inference (dotted arrows) for improved detection. The decoder learns to associate the derived exemplar embeddings with the characteristics of the class in an image, along with the text, thus allowing minimal-shot detection. This approach lays the foundation for a continuous learning framework with minimal annotations when new findings are introduced.}
    \label{fig:overview_graphical_abstracts}
\end{figure}
\end{graphicalabstract}

\begin{highlights}
\item Scaling Exemplar Med-DETR with Minimal Annotations: Demonstrates how exemplar-based detection can effectively scale to large chest radiography datasets with minimal annotation effort, reducing reliance on extensive labeled data for new findings.
\item Adaptable Framework for Emerging Findings: Proposes a structure that could be extended to accommodate new disease classes with limited retraining, offering a foundation for future continuous learning that avoids catastrophic forgetting on previously learned classes.
\item Includes evaluations on both proprietary and public CXR datasets with ablation studies, demonstrating robustness and scalability under minimal-annotation and out-of-distribution settings.
\end{highlights}


\end{frontmatter}
\section{Introduction}
Deep learning has achieved remarkable progress in visual understanding tasks, yet these advances are heavily dependent on large-scale annotated datasets \cite{lin2014microsoft,everingham2010pascal}. In the medical imaging domain, obtaining such annotations remains a persistent bottleneck, as each label requires expert interpretation that is time-consuming, and often varies across institutions and modalities \cite{emdetr,nguyen2022vindr,aljabri2022towards}. Reducing the annotation effort while maintaining high diagnostic performance has therefore been a central research objective for Computer-Aided Diagnostic (CAD) systems \cite{pujitha2018solution,monkam2023annotation}. 

In pursuit of this goal, various strategies have been proposed in the literature aimed at reducing the dependence on manual expert annotations. These approaches range from weakly-supervised and semi-supervised learning methods that exploit incomplete labels \cite{jang2020deep,choi2022semi,bhat2025patch}, as well as active learning methods that selectively query the most informative samples\cite{pachetti2024systematic}, and few-shot learning (FSL) methods that aim to generalize from only a handful of labeled examples per class \cite{bulat2023fs,zhang2022meta}. Nevertheless, these methods have not yet demonstrated competitive spatial precision at scale in medical imaging tasks \cite{müller2024weaklysupervisedobjectdetection}. 

In contrast to these annotation-efficient approaches, recent multi-modal Vision-language (VL) detection methods \cite{liu2024grounding,li2022grounded,minderer2022simple} demonstrate zero-shot, open-vocabulary detection capabilities over a wide variety of target classes in computer vision datasets \cite{lin2014microsoft,everingham2010pascal}. These results are achieved in part through strong pretraining and the use of a large number of semi-automatically generated ``grounding'' captions. 
However, transferring these methods to the medical domain remains difficult: radiology reports rarely provide explicit fine-grained grounding descriptions, and similar large-scale grounded medical datasets are not available. As a result, integrating annotations with descriptive text and spatial location, despite its potential to reduce the burden of annotations, remains largely underexplored in medical imaging.

In Chest X-Ray (CXR) research, current state-of-the-art (SOTA) systems rely extensively on both increased number of annotations per finding and post-processing techniques such as handcrafted Non-Maximal Suppression (NMS) \cite{ghesu2022contrastive} methods to achieve high diagnostic accuracy. Nonetheless, in the current age of automated report production \cite{sloan2024automated,yu2023evaluating,liu2019clinically}, there is an increasing need for CAD systems that can detect findings with minimal annotations. This need is particularly evident in CXR datasets, which often contain long-tailed label distributions in which many findings are represented by only a small number of samples \cite{nguyen2022vindr,lin2025cxr}. Thus, it is beneficial to develop a CAD system that learns from \emph{base class(es)} where training data are inexpensive to acquire, available in abundance; and therefore can achieve optimal performance with minimal annotations for novel or rare findings. 

These requirements motivated our example-based abnormality detection method, called Exemplar Med-DETR (EM-DETR) \cite{emdetr}, which reduces the dependence on dense annotations and avoids post-processing. Although inspired by VL detection architectures, EM-DETR operates on learnt visual exemplars, making it suitable for domains without grounded radiology captions. In our previous work, we demonstrated that EM-DETR achieves SOTA performance in multiple medical domains, such as the detection of mass and calcification in mammography, masses in CXR images, and stenoses in angiography images \cite{emdetr}. In this study, we evaluate the performance of EM-DETR by simulating a minimal-shot scenario with a set of novel or rare classes that are held out during base class training. This allows us to assess the ability of EM-DETR to detect clinically important but sparsely annotated findings under realistic annotation constraints.

With the recent success of transformer-based multi-modal detection pipe\-lines, our proposed approach is based on the principles of grounding DINO \cite{liu2024grounding}, which extends GLIP \cite{li2022grounded} and DETR \cite{carion2020end} to tightly fuse language and vision modalities through a feature enhancer and cross-modality decoder. To incorporate prototypical visual cues, we introduce an Exemplar Generation Module $\mathcal{G}$, which serves as an additional input to the decoder (see Fig. \ref{fig:pretraining_overview}). The module derives an exemplar or prototype class-specific embedding for each class. This embedding is included in the cross-attention pipeline such that the decoder detects regions that not only match the text description but also the visual features of that class. In other words, to the existing multi-modal framework, we add class-conditional knowledge (few-shot exemplars or prototypes) into the detection process. This encourages decision boundaries that reflect semantic distances between findings rather than relying solely on their correlation with text descriptions. Theoretically, this enables improved generalization to unseen domains or novel classes, supports few-shot learning through stable prototype anchors, and reduces overfitting to frequently occurring classes.

In addition to architectural modifications, we also adapt the model to the unique characteristics of the CXR data, which differ substantially from natural image datasets. As CXR images are acquired under standardized protocols,  they exhibit strong spatial alignment \cite{tafti2020x}, with several abnormalities consistently appearing in specific anatomical regions. 
Furthermore, unlike natural images where class features are typically distinct (e.g., car vs. tree), CXR findings often exhibit high intra-class variability due to disease manifestation \cite{galan2024few}, and the overlapping features of different abnormalities.  Accurate diagnosis also depends on contextual, often non-imaging, information \cite{9395510}, increasing the difficulty of purely appearance-based discrimination. Given these domain-specific challenges and the reliance of EM-DETR on contrasting classes, we find that explicit selection of negative samples plays a critical role in enhancing localization accuracy—even in minimal-annotation regimes.

Our main contributions are summarized as follows:
\begin{itemize}
    \item \textbf{Scaling EM-DETR with minimal annotations:} Demonstrates how exemplar-based detection can scale effectively in chest radiographs using only a small number of labeled samples, reducing annotation burden for clinical deployment.
    
    \item \textbf{Adaptable framework for new findings:} Proposes an architecture designed to facilitate addition of new disease classes with limited retraining, forming a basis for future work in continuous learning.
    
    \item \textbf{Domain-aware iterative negative sampling:} Introduces a domain-specific background selection strategy for iterative contrastive learning, improving training stability and detection accuracy in multi-class, long-tailed scenarios.
    
    \item \textbf{Evaluation on CXR datasets:} Validated across proprietary and public datasets with ablation studies, including an out-of-distribution test, demonstrating robustness under minimal-annotation scenarios.
    
\end{itemize}



\section{Related Works}

The field of abnormality detection in radiology has been extensively researched in the domain of Artificial Intelligence (AI), with numerous SOTA techniques available that include open-source \cite{9363915,müller2024weaklysupervisedobjectdetection,lin2022cxr} and commercial options \cite{9395510,9098551,lunit2020insight,qure2020validation,putha2018can,huang2022annalise}. Approaches to reduce manual annotations for AI training is one of the most active research areas in medical AI at the moment \cite{galan2024few,zheng2024cross,lu2023medoptnet,moor2023med}. These strategies can be broadly categorized into weakly- or semi-supervised learning, multi-modal learning, and few-shot learning approaches, spanning tasks in classification, detection, and segmentation across both computer vision and medical domains. In the following sections, we review the current research landscape in these domains, highlighting recent advances and methodological trends. 


\subsection{\textbf{Weakly or semi-supervised methods}}

Weakly or semi-supervised frameworks leverage unlabelled or weakly labelled data through pseudo-labels obtained by various methods such as clustering, consistency regularization or contextual label generation. Recent studies by Ju et al. \cite{ju2024universal} address universal semi-supervised learning for classification by employing a variational auto-encoder (VAE). The VAE is pretrained to understand unseen samples based on the assumption that these samples follow a different distribution. The authors propose a unified framework using a "dual-path outlier estimation scheme". While elegant, this approach is strongly dependent on the Gaussian distribution assumption of the underlying data. Alternatively, He et al. \cite{he2024open} demonstrate open-set semi-supervised classification through a multi-binary discriminator and a joint outlier filter mechanism, learning prototypes that help identify distinct classes. 

In the field of segmentation, Du et al. \cite{liao2024weakly} propose using a geometric prior along with contrastive loss to improve 3D segmentation with only bounding box annotations and Kuang et al. \cite{kuang2024weakly} propose multi-class segmentation through ``feature decomposition'. Likewise, Li et al. \cite{li2023weakly} present a weakly supervised segmentation approach for histopathology images using multi-instance learning (MIL) with self-attention, effectively aggregating global contextual information through deep supervision. Alternatively, the authors of INSIGHT \cite{zhang2024insight} draw on general vision strategies, combining a detection module with a heatmap aggregator for classification and segmentation at the same time. For a broader overview, Jiao et al. \cite{jiao2024learning} provide a comprehensive survey of semi-supervised learning techniques for medical image segmentation. 

Beyond the medical domain, semi-supervised and weakly supervised methods have also been extensively explored in general computer vision tasks. Weak supervision in computer vision typically uses box, point annotations \cite{zhang2022group} and image-level labels, while annotation types in medical imaging employ points, scribbles, and gaze \cite{wang2025improving,chen2025gaze}. Moreover, in the medical domain, there is a greater influence of domain shift, spatial resolution, and class imbalance compared to the general image domain. Recent studies in the latter domain include STAC \cite{sohn2020simple} and object detection via teacher-student training \cite{choi2022semi,xu2021end}, which are used to produce pseudo-labels for unlabeled images. On the other hand, for holistic scene understanding, weakly and semi-supervised panoptic segmentation methods have emerged as efficient alternatives to full supervision. Works such as Li et al. \cite{li2024complete}, Li et al. \cite{Li_2018_ECCV}, and Lee et al. \cite{Lee_2021_ICCV} integrate semantic and instance cues through proposal refinement, region-to-pixel consistency, and cross-task distillation, respectively, to jointly segment “thing” and “stuff” categories with limited labels. Nevertheless, such progress does not directly translate to medical detection in CXRs, where the challenge lies in localizing subtle findings rather than performing generic classification or pixel-level segmentation. In the following section, we investigate multi-modal approaches that facilitate DL tasks with minimal annotations.

\subsection{\textbf{Vision-Language (VL) methods}}

Recent multimodal methods, especially vision–language frameworks, improve model performance by integrating contextual knowledge from accompanying text, effectively using language as a weak supervisory signal. Early studies, including UNITER \cite{chen2019uniter} and OSCAR \cite{nguyen2024oscar}, focus on developing universal embeddings or features by pretraining with image-caption datasets in the natural image domain. UNITER \cite{chen2019uniter} presents four pretraining tasks, including masked language/region modeling, VL matching, and optimal-transport word–region alignment, aimed at achieving detailed text and image region alignment. OSCAR \cite{nguyen2024oscar} similarly uses object tags as “anchor” tokens to align image regions with words, pretraining on 6.5M image–text pairs and achieving SOTA on VQA, retrieval, and grounding benchmarks \cite{Li_2025_CVPR,singh2019towards,liu2024mmbench,reiter2018structured,yu2023mm}. These methods illustrate the core paradigm of concatenating region features (from object detectors) with text tokens and applying cross-modal self-attention. In this context, the text acts as a distinctive encoding key, enabling the transformer encoder-decoder mechanism to effectively recognize image data from the provided textual input.

Likewise, recent methods like GLIP\cite{li2022grounded} and OWL-ViT\cite{minderer2022simple} develop open-vocabulary detectors that utilize contrastive image-text training with vision transformers, enabling text-conditioned training on arbitrary categories.  The most recent VL model Grounding DINO \cite{liu2024grounding} further improves upon GLIP\cite{li2022grounded} by combining the powerful DINO backbone with grounded language-based object detection, leading to further enhancements in zero-shot object detection performance. This approach incorporates enhancements to the Detection Transformer Encoder-Decoder (DETR), including the use of a deformable MS-attention encoder along with robust pretraining utilizing captions and descriptions from over 1.8 million images. This setup facilitates open-set detection for wide variety of unseen classes. On the other hand, BLIP \cite{li2022blip} focuses on data: It bootstraps noisy web images by using a captioner to generate high-quality captions and a filter to remove noise, enabling unified pretraining for both understanding and generation. BLIP achieves strong gains in retrieval, captioning, and VQA by leveraging this synthetic web-scale supervision.

Beyond alignment and detection, large multi-modal methods now handle open-ended and few-shot tasks. Flamingo \cite{alayrac2022flamingo} is a few-shot VLM that interleaves pretrained vision and language modules via a “perceiver” resampler, later extended to Med-Flamingo \cite{moor2023med}. The model processes mixed text and images to address open-ended tasks through prompts. With only a handful of examples, Med‑Flamingo shows SOTA few‑shot performance on medical VQA benchmarks, outperforming methods that were fine‑tuned on much larger datasets. 
Analogously, Kosmos-2 \cite{peng2023kosmos} extends these ideas to a generalist multi-modal LLM. Trained from scratch on web-scale interleaved image–text data, Kosmos-2 \cite{peng2023kosmos} supports zero-/few-shot learning, instruction following, and even OCR-free language tasks by converting visual inputs into “language” tokens. It shows strong performance on various VL benchmarks such as captioning and VQA. In contrast, MedOptNet \cite{lu2023medoptnet} proposes a meta-learning framework utilizing high-performance convex optimization methods for few-shot classification. 


Despite impressive results on few-shot classification and reporting tasks, the medical domain lacks large-scale grounding text paired with bounding box annotations, limiting the applicability of these methods for few-shot detection. VL methods can reduce annotation needs, but only when textual cues provide meaningful guidance. Methods such as Grounding DINO \cite{liu2024grounding} under-perform on unseen classes in few-shot evaluations, and Flamingo \cite{alayrac2022flamingo} struggles to localize specific findings. Language-based supervision is further limited in the medical domain due to scarce pretraining data, a constrained and non-standardized vocabulary, and frequent overlap of terms across different findings \cite{pmlr-v227-windsor24a,Sha_Fewshot_MICCAI2024,pachetti2024systematic}.

\subsection{\textbf{Few-shot learning methods}}

Few-shot detection enables models to adapt to new classes with minimal samples, leveraging pretraining on base classes. Bulat et al. \cite{bulat2023fs} introduce FS-DETR, a few-shot detection transformer that uses visual prompts to detect novel classes. By generating visual templates and pseudo-class embeddings with a strong image backbone, FS-DETR outperforms zero-shot and even some fine-tuning methods on benchmarks like PASCALVOC \cite{everingham2010pascal} and MSCOCO \cite{lin2014microsoft}. In natural images, ROIs (e.g., “cat” or “dog”) yield discriminative embeddings from ImageNet-pretrained CNNs \cite{deng2009imagenet}, enabling effective template matching. In contrast, CXR ROIs such as “pleural effusion” or “lesion” produce overlapping embeddings \cite{Cohen2022xrv} (see Fig. \ref{fig:classifier_tsne}, \ref{app_1}), influenced by anatomical structures and lacking isolated class examples. As a result, the decoder must distinguish pathological from normal anatomy using both global and local context, motivating the need for a robust template generator that produces highly discriminative, class-specific features.

Meanwhile Zhang et al.\ present Meta-DETR \cite{zhang2022meta}, that bypasses region proposals and uses an inter-class correlational meta-learning strategy to improve few-shot object detection, mainly achieved by capturing the correlation between classes. These methods highlight the trend toward encoder-decoder architectures that treat all V+L tasks as generative language modeling. While both FS-DETR \cite{bulat2023fs} and Meta-DETR \cite{zhang2022meta} extend DETR for few-shot detection using meta-learning concepts, FS-DETR \cite{bulat2023fs} primarily focuses on feature re-adaptation and query re-weighting, whereas Meta-DETR introduces a meta-learning module that learns transferable parameters for task-level adaptation. Although effective on computer vision datasets, these methods do not translate as strongly across all CXR medical findings. 

Complementing these studies, several recent works have advanced few-shot image classification through diverse representation learning strategies. 
Zheng et al. \cite{zheng2023unsupervised} proposed Unsupervised Few-Shot Image Classification via One-vs-All Contrastive Learning, extending few-shot learning to unsupervised settings by leveraging contrastive objectives. Building on these foundations, Semantic-Guided Generalization Enhancement (SGE) \cite{zheng2025sge} integrates semantic priors to guide generalization, addressing the semantic shift across domains in few-shot learning. Similarly, Wu et al. \cite{wu2023invariant} developed invariant and consistent unsupervised representation learning for few-shot visual recognition, emphasizing stability under feature transformations. Together, these studies illustrate a coherent research trajectory focused on enhancing feature robustness, semantic consistency, and generalization in low-data regimes, forming a solid foundation for subsequent multimodal or domain-adaptive extensions such as MERGE \cite{zheng2025merge} and DARA \cite{wu2025dara}. 

Converse to previous work focussing on classification, Wang et al. presented a prototype-based feature mapping network designed to address the adaptation of the few-shot domain in medical segmentation in PFMNet \cite{wang2024pfmnet}. However, these approaches are primarily designed for natural image domains within the few-shot learning paradigm, often focus solely on classification or segmentation, and do not readily translate to robust few-shot detection of multiple findings in CXR medical imaging. Furthermore, open-source implementations are limited, hindering replication and extension. Building on these observations, we next detail the dataset preparation and the proposed EM-DETR adaptation for minimal-shot CXR detection, before presenting the experimental results.

 
\section{\textbf{Data preparation}}
We first examine our hypothesis on a smaller anatomy dataset to verify the feasibility of few-shot detection using EM-DETR. This approach is logical, as anatomy detection on CXRs is less challenging compared to the detection of complex findings. Subsequently, we proceed to assess EM-DETR using minimal annotations on larger standard datasets and intricate findings.

Towards this end, two separate datasets ($\mathcal{A}$ and $\mathcal{B}$) are derived from a larger private dataset that was acquired from multiple sites in the United States and Europe. The study includes frontal and lateral chest exams from adult cohorts ($>$18 years). Two general radiologists independently reviewed and annotated all CXRs with the presence and location of 16 findings. A third general radiologist served as an independent tiebreaker for discrepant findings. The following subsections detail the private dataset specifics, and then describe the public dataset used for further evaluation.

\subsection{\textbf{Proprietary CXR Anatomy dataset}}
\begin{figure}
    \centering
    \includegraphics[width=0.8\linewidth]{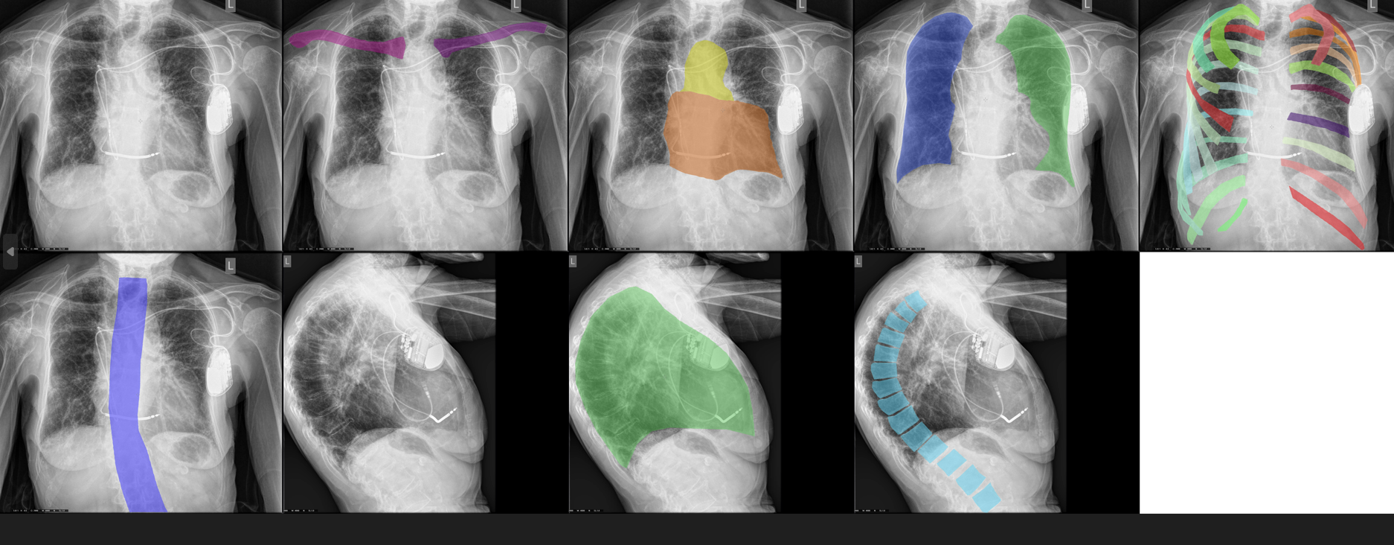}
    \caption{Example image from the anatomy dataset, where we only use the frontal views and discard the lateral views.}
    \label{fig:anatomy_sample}
\end{figure}

We use a dataset $\mathcal{A}$ of anatomy annotations, comprising 100 full-resolution anonymized DICOM images. The dataset includes 80 frontal images for training and 20 for validation. The splits are obtained patient-wise to avoid patient overlap. The images contain segmentation masks for 7 of the major organs such as the lungs, clavicle, heart, aorta, and ribs, as shown in Fig. \ref{fig:anatomy_sample}.  

The 7 segmentation masks are first converted into bounding box coordinates. These annotations are further separated into left and right organ classes, resulting in 10 classes. Table. \ref{tab:anatomy_detection} denotes the final number of classes used to train EM-DETR. Moreover, the ribs and spine annotations are also divided into 3 separate classes: upper, middle, and lower organ. The demarcations are obtained using a third of the original height of the bounding box. To simulate a few-shot condition, only five training images include annotations of the heart region. 

\subsection{\textbf{Proprietary CXR Abnormality dataset}}
A second dataset $\mathcal{B}$ of 18,681 full-resolution anonymized DICOM CXR images is used to evaluate the few-shot performance of EM-DETR on 3 different CXR findings. The dataset includes frontal-view images of patients, annotated with bounding boxes indicating the presence of 16 findings provided by an expert panel of radiologists. We use a split of 3 validation datasets of 331, 412 and 412 images for pleural effusion, pneumothorax (ptx) and lesion detection. The validation datasets have binary labels \{1,0\} indicating the existence or absence of a finding. 
We ensure that there is no overlap between the patients in the training and validation splits and Table.~\ref{tab:data_split} presents the data split with the number of positive samples in brackets. 

\begin{table}[h]
\footnotesize
    \centering
    \begin{tabular}{lcc}
        \toprule
      \multirow{2}{*}{\textbf{Finding}} & \multicolumn{2}{c}{\textbf{Images (Positive samples)}} \\
        \cline{2-3}
         & \textbf{Training} & \textbf{Test} \\
        \midrule
        Pleural effusion & 18,681(4037) & 331(74) \\
        Pneumothorax & 18,681(1868) & 412(146) \\
        Lesion & 18,681(5654) & 412(109) \\
        \bottomrule
    \end{tabular}
    \caption{Number of images per finding in the training and test sets. The number in the brackets indicate the number of positive samples for each finding.}
    \label{tab:data_split}
\end{table}

Pleural effusion is a relatively easy finding for the network to detect while pneumothorax and lesion are more challenging. Together, they cover the spectrum of difficulty for detection in CXRs \cite{bhat2023aucreshaping,bhat2025patch}. 
To simulate a minimal-shot scenario, a pretraining dataset called $\mathcal{B}_P$ is created with all findings excluding pleural effusion, pneumothorax, and lesion. The finetuning dataset $\mathcal{B}_F$ is then composed of a smaller number of images with the finding of interest and a sufficient number of images without the finding, which are taken as negative samples. For eg.\ a smaller subset of 200 images with pneumothorax and 800 images without pneumothorax is taken as the finetuning dataset for pneumothorax. This accounts for 7\% of the entire training dataset, where the samples are selected at random. This training scheme simulates a minimal-shot scenario, in which we have a powerful pretrained model, and introduce a new class with minimal annotations. 

\subsection{\textbf{Public CXR Abnormality dataset}}

Additionally, we test the finetuned EM-DETR model on the public VinDR-CXR test dataset \cite{nguyen2022vindr} without any further finetuning. The VinDR-CXR \cite{nguyen2022vindr} testing dataset comprises 3000 images and features 22 local labels alongside 6 global labels. We test the classification and FROC performance for pneumothorax due to it being a difficult task with only 18 positive samples in this test dataset.


\section{Method}

\begin{figure}
    \centering
    \includegraphics[width=1\linewidth]{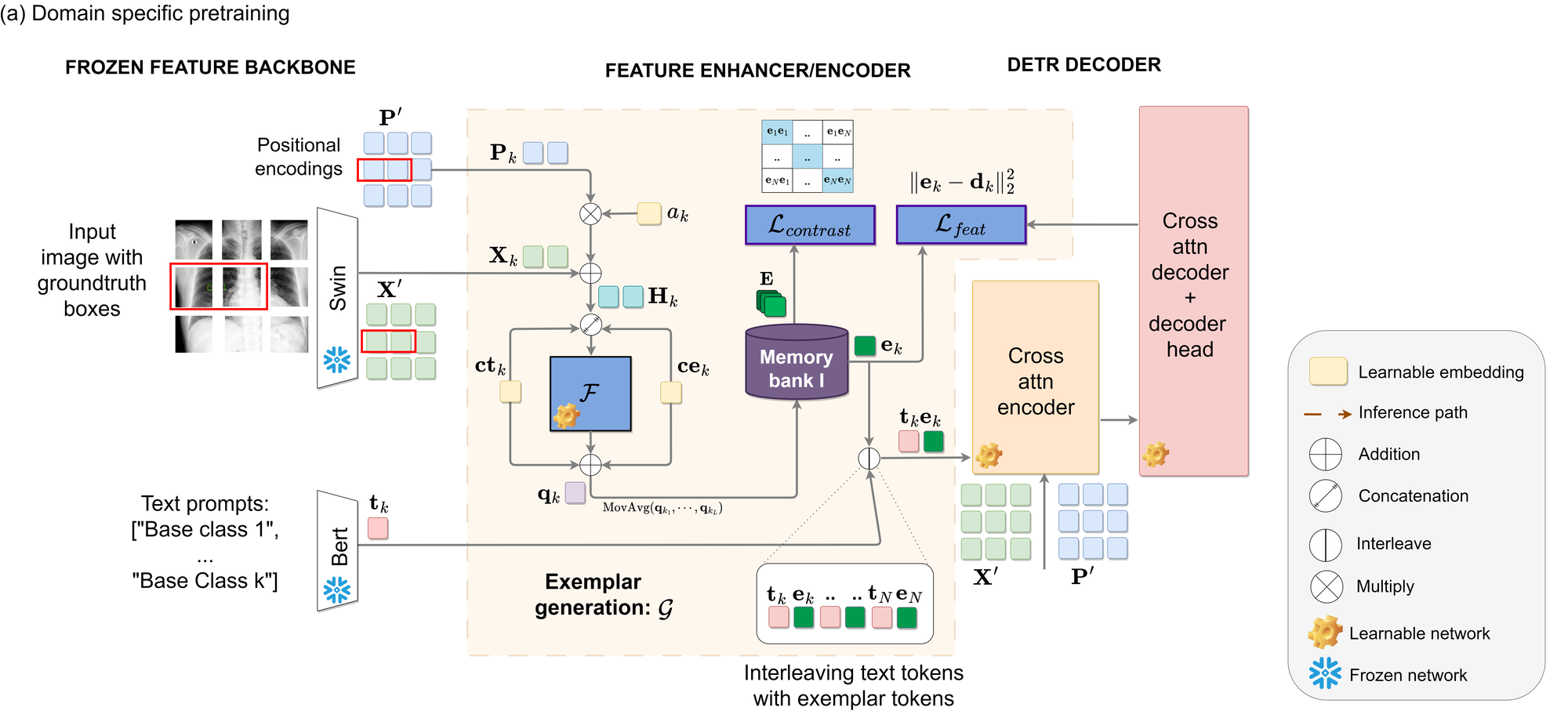}
    \caption{Overview of EM-DETR pretraining. Visual features $\mathbf{X}_k$ \& positional embeddings $\mathbf{P}_k$ are extracted from the frozen image backbone based on the $k^{th}$ class location (red). $\mathbf{F}$ uses learnable class-specific embeddings $\mathbf{ct}_k$ \& $\mathbf{ce}_k$ to calculate $\mathbf{q}_k$. A moving average of $\mathbf{q}_k$ results in the prototype embedding $\mathbf{e}_k$. Interleaving the text embedding $\mathbf{t}_k$ with $\mathbf{e}_k$ enables text-and feature-based detection for each base class. $\mathbf{e}_k$ is used in additional $\mathcal{L}_{contrast}$ \& $\mathcal{L}_{feat}$ losses to improve the decoder search operations.}
    \label{fig:pretraining_overview}
\end{figure}

Inspired by FS-DETR \cite{bulat2023fs}, we adapt our previously introduced EM-DETR \cite{emdetr} for minimal-shot detection. The proposed framework consists of two main components: exemplar-based finetuning and iterative contrastive training. In the following subsections, we first describe how per-finding exemplars are generated to enhance detection, and then explain the iterative training scheme.

EM-DETR builds on multi-modal DETR \cite{carion2020end,liu2024grounding}, performing ``language-guided query selection'' via cross-attention between image and text embeddings. To guide detection, we learn class-specific exemplars computed from visual features corresponding to each class’s spatial location, bootstrapping template generation similarly to FS-DETR. A moving average of these exemplars are interleaved with the text embeddings and passed to the Detection Transformer Encoder-Decoder pipeline. Attending to these exemplars enables the detection heads to perform prototype-based searches, facilitating straightforward expansion to novel classes. Fig. \ref{fig:pretraining_overview} illustrates the EM-DETR pretraining stage, including exemplar generation and the additional losses that regularize the generated exemplars.

To further improve learning, we adopt an iterative contrastive strategy. In each stage, hard negatives are mined either by generating background annotations from normal images or by leveraging false positives, refining class prototypes and improving overall detection performance.

\subsection{\textbf{Exemplar Generation $\mathcal{G}$:}}\label{eg_subsection}
The Swin transformer \cite{liu2021swin} image backbone is a multi-scale, shifted window transformer that produces a set of $J$ patch embeddings $\mathbf{X}'=[\mathbf{x_1} 
,... \mathbf{x}_J]^T$ for an input image, where each $\mathbf{x}_j \in \mathbb{R}^d$. A similar dimension set of positional encodings is also generated, denoted as $\mathbf{P}'=[\mathbf{p_1},... \mathbf{p}_J]^T$. Let $\mathbf{X}_k \subset \mathbf{X}'$ and $\mathbf{P}_k \subset \mathbf{P}'$ represent the set of $M$ tokens that fall within the ROI derived from the bounding box ground truth annotations of the class  $k$ (red). 

In CXRs, certain findings consistently manifest in predictable patterns and are frequently localized to specific areas, such as in the cases of ``pleural effusion'' and ``cardiomegaly''. Conversely, other findings exhibit significant variability in both presentation and anatomical location. To account for this, $\mathbf{P}_k$ is scaled by a learnable class-specific parameter $a_k \in \mathbb{R}$ and added to $\mathbf{X}_k$. This produces $\mathbf{H}_k$, mathematically shown as
\begin{align}
\label{eq:1}
    \mathbf{H}_k &= \mathbf{X}_k + \mathbf{P}_k \times a_k.
\end{align}
$a_k$ modulates the impact of the positional embeddings, thereby influencing the decoder search operation. 
Additional learnable class-wise token $\mathbf{ct}_k$ and positional c embeddings  are then concatenated with $\mathbf{H}_k$. The simple attention pooling transformer $\mathcal{F}$  processes the concatenated result to learn a single normalized embedding using self-attention mechanism \cite{marin2023token}. The use of $\mathbf{ct}_k$ and $\mathbf{ct}_k$ within $\mathcal{F}$ provides additional class-specific information that helps to build a unique example feature. Concurrently, $\mathbf{ct}_k$ and  $\mathbf{ce}_k$ are also added to the output of $\mathcal{F}$ to produce $\mathbf{q}_k \in \mathbb{R}^d$ encapsulating the textural and spatial features corresponding to class $k$. The addition of $\mathbf{ct}_k$ and $\mathbf{ce}_k$ is analogous to the positional embeddings added to the text embeddings in the feature enhancer \cite{liu2024grounding,zhang2022meta}. 
Mathematically, $\mathbf{q}_k$ is given as 
\begin{align}
\label{eq:2}
    \mathbf{q}_k &=\mathcal{F}(concat(\mathbf{ct}_k, \mathbf{ce}_k, \mathbf{H}_k)) + \mathbf{ct}_k + \mathbf{ce}_k,  
\end{align}
where $k \in \{1, \dots, N\}$ and $N$ is the total number of classes. A per-class moving average is calculated over $L$ samples of $\mathbf{q}_k$, thereby generating the prototype embedding $\mathbf{e}_k \in \mathbb{R}^d$. This prevents the decoder from pursuing rapidly changing features during each training iteration. $\mathbf{e}_k$ is stored in a \emph{Memory bank} as a prototype feature embedding that helps prevent catastrophic forgetting. Eventually, model inference relies on the memory bank of saved feature representations, or exemplars $\mathbf{e}_k$, since ROI regions are not available for $\mathbf{e}_k$ generation in test images.

\begin{figure}
    \centering
    \includegraphics[width=0.6\linewidth]{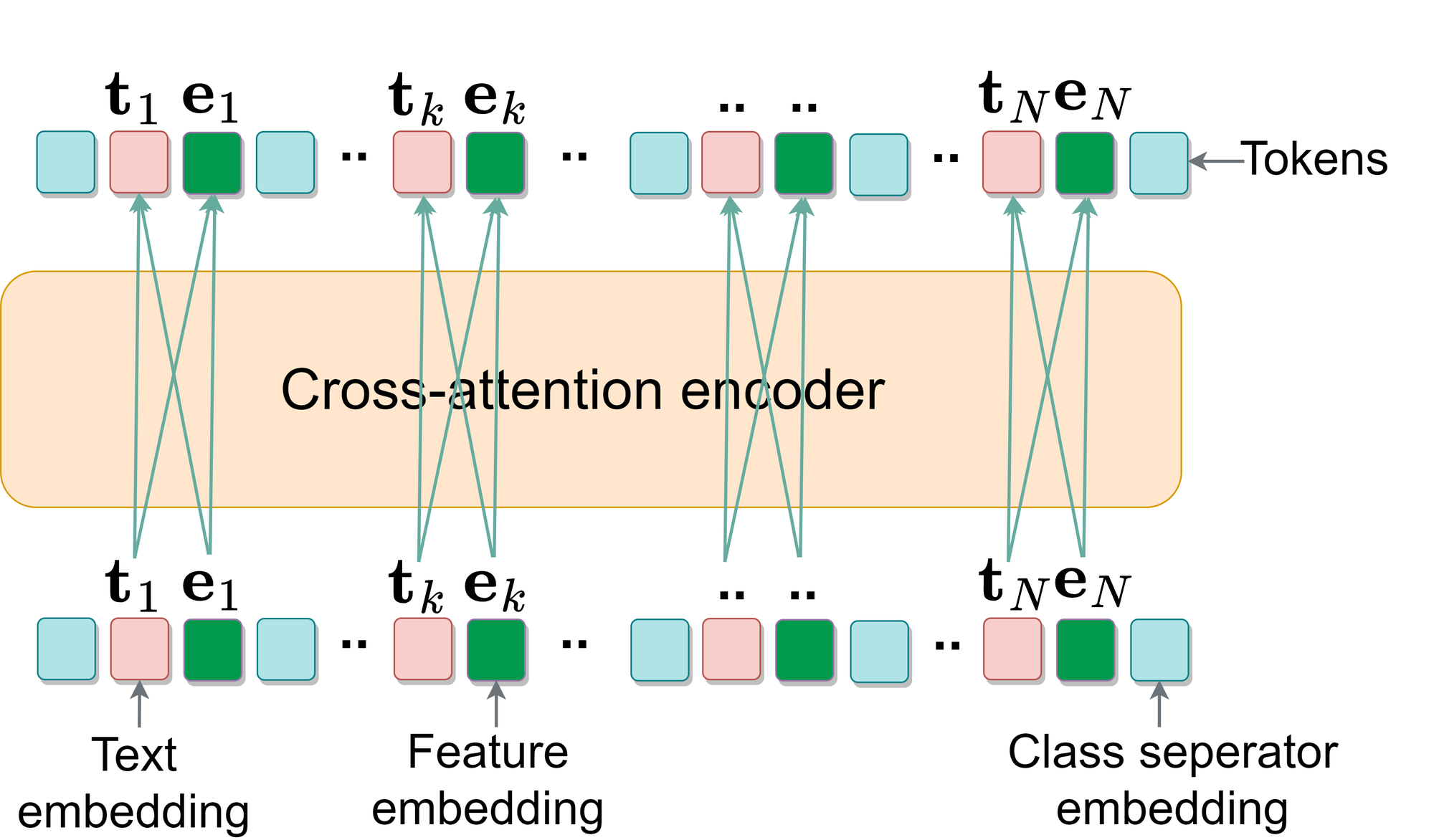}
    \caption{Representation of the per-class interleaving of the text and feature embeddings. The tokens generated from the encoder undergo self-attention at a sub-sentence level through additional attention masks \cite{liu2024grounding}. }
    \label{fig:interleave}
\end{figure}

In parallel, the frozen text encoder processes text prompts and produces embeddings $\mathbf{t}_k$ which are interleaved with the corresponding $\mathbf{e}_k$ and passed downstream to the encoder-decoder pipeline \cite{carion2020end,liu2024grounding}. The text prompt inputs are literal class names (e.g. ``pleural effusion'', ``background''). Fig. \ref{fig:interleave} illustrates the interleaving and encoding operations in greater detail. Attention masks ensure that self-attention is applied within the tokens for each class at a sub-sentence level similar to \cite{liu2024grounding}.

Thereafter, the decoder predicts the location of the $k^{th}$ class by processing the cross-attention encodings of the text and representative visual features, $\mathbf{t}_k$ and $\mathbf{e}_k$ with the input image embeddings, $\mathbf{X'}$ and $\mathbf{P'}$. This is done through a contrastive search that looks for highest cosine similarity between the many hypothesis boxes and the text or feature embeddings. 

\paragraph{\textbf{Additional Losses $\mathcal{L}$:}}\label{loss_subsection} In addition to the original DETR loss terms $\mathcal{L}_{\text{bbox}}$, $\mathcal{L}_{\text{IoU}}$, and $\mathcal{L}_{\text{classify}}$ in \cite{liu2024grounding} we introduce two additional loss functions to improve the robustness of EM-DETR. A cosine similarity \emph{contrastive feature loss} $\mathcal{L}_{contrast}$, is applied on $E=[\mathbf{e}_1...\mathbf{e}_N]^T$ of all $1\le k \le N$ classes.  $\mathcal{L}_{contrast}$ ensures that all class prototype embeddings remain orthogonal to each other in the latent space, promoting distinct and separable representations \cite{yang2022class}. 
A \emph{$L_2$ feature loss}, $\mathcal{L}_{feat}$, is applied between $\mathbf{e}_k$ and the decoder's top proposal $\mathbf{d}_k$ for class $k$ ensuring that the model predicts a consistent latent representation for each class \cite{bulat2023fs,zhang2022meta}. This approach helps maintain class-specific embeddings over time while stabilizing the training process. Although previous studies\cite{bulat2023fs,zhang2022meta} indicate that $\mathcal{L}_{feat}$ does not significantly affect precision results in general vision studies, it is empirically observed to improve our Free Response Operating Characteristic (FROC) results \cite{emdetr}.

\subsection{\textbf{Iterative training strategy:}}\label{iterative_subsection}
\begin{figure}
    \centering
    \includegraphics[width=0.6\linewidth]{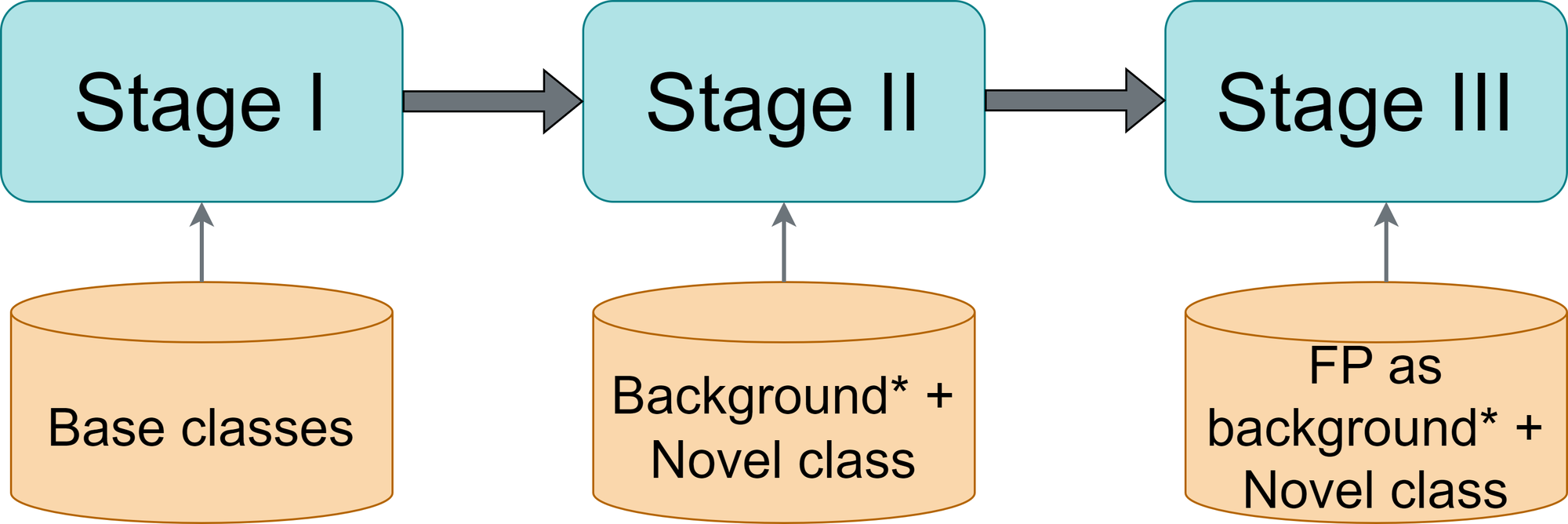}
    \caption{Schematic of iterative training strategy, where $*$ denotes generated annotations.}
    \label{fig:iterative_learning}
\end{figure}

The trained decoder learns class-specific features but struggles to distinguish normal anatomy from disease traits in abnormality detection. Therefore, we propose a multi-stage iterative learning approach, for effective minimal-shot abnormality detection. Fig.~\ref{fig:iterative_learning} illustrates the iterative training schematic. Stage I involves pretraining the proposed model with all base class annotations dataset $B_P$, while subsequent finetuning stages refine the weights through a binary per-class background-versus-foreground detection task using only the minimal positive samples. 

\begin{figure}
    \centering
    \includegraphics[width=0.85\linewidth]{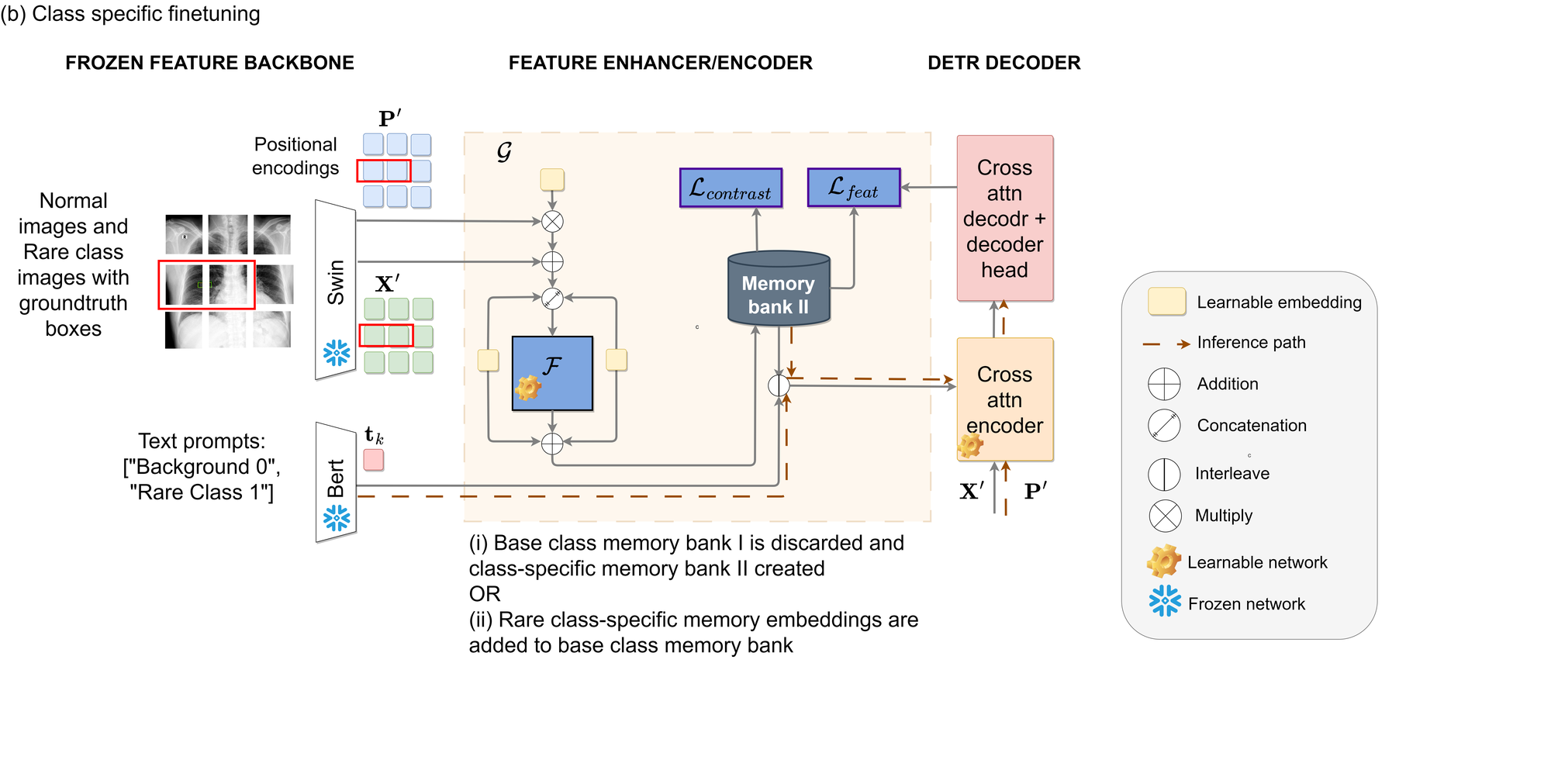}
    \caption{Overview of the finetuning and inference process. The memory bank accumulated during Stage I is either discarded or the novel class embeddings are added to create a new memory bank II. The inference path is shown in dotted arrow line.}
    \label{fig:finetuning_overview}
\end{figure}

Fig.~\ref{fig:finetuning_overview} illustrates the EM-DETR workflow for finetuning and inference. In our framework, finetuning is not restricted to a single stage but is designed as a modular, multi-stage strategy that incrementally refines the model for a given novel class. For example, Stage II initiates this process, and additional stages can be appended to further improve performance by leveraging feedback from prior stages, such as false positive suppression. This design allows continued adaptation without retraining from scratch. 

In Stage II training, the pretrained model is trained on dataset $B_F$. The pretrained memory bank-I of the base classes, can either be discarded to create a binary class memory bank-II or extended by incorporating novel class exemplars, depending on the model design. Since the network's classification layer aligns text prompts with hypothesis regions, freezing the decoder doesn't significantly impact performance. However, freezing the encoder in Stage II impairs performance, as the novel class features require cross-attention encoding. $\mathcal{F}$ stays trainable to ensure effective attention pooling for the novel classes. 

In an additional Stage III, we evaluate the Stage II model on the training dataset and denote the False Positive (FP) regions as background classes to further refine the network. This helps the model refine its ability to separate visually ambiguous areas from actual pathological findings. During inference, the memory bank is used exclusively, as there are no ROIs available.

\paragraph{\textbf{Background generation:}}\label{background_subsection}

The effectiveness of the contrastive detection model depends on the contrastive learning scheme between classes. In medical applications such as CXR, two main challenges arise: differentiating anatomical features from abnormalities and distinguishing closely related class features. To address this, we reformulate the multi-class problem into a binary detection task by negative mining the background class from images that do not contain the finding. 

\begin{figure}[ht]
    \centering
    \includegraphics[width=0.7\linewidth]{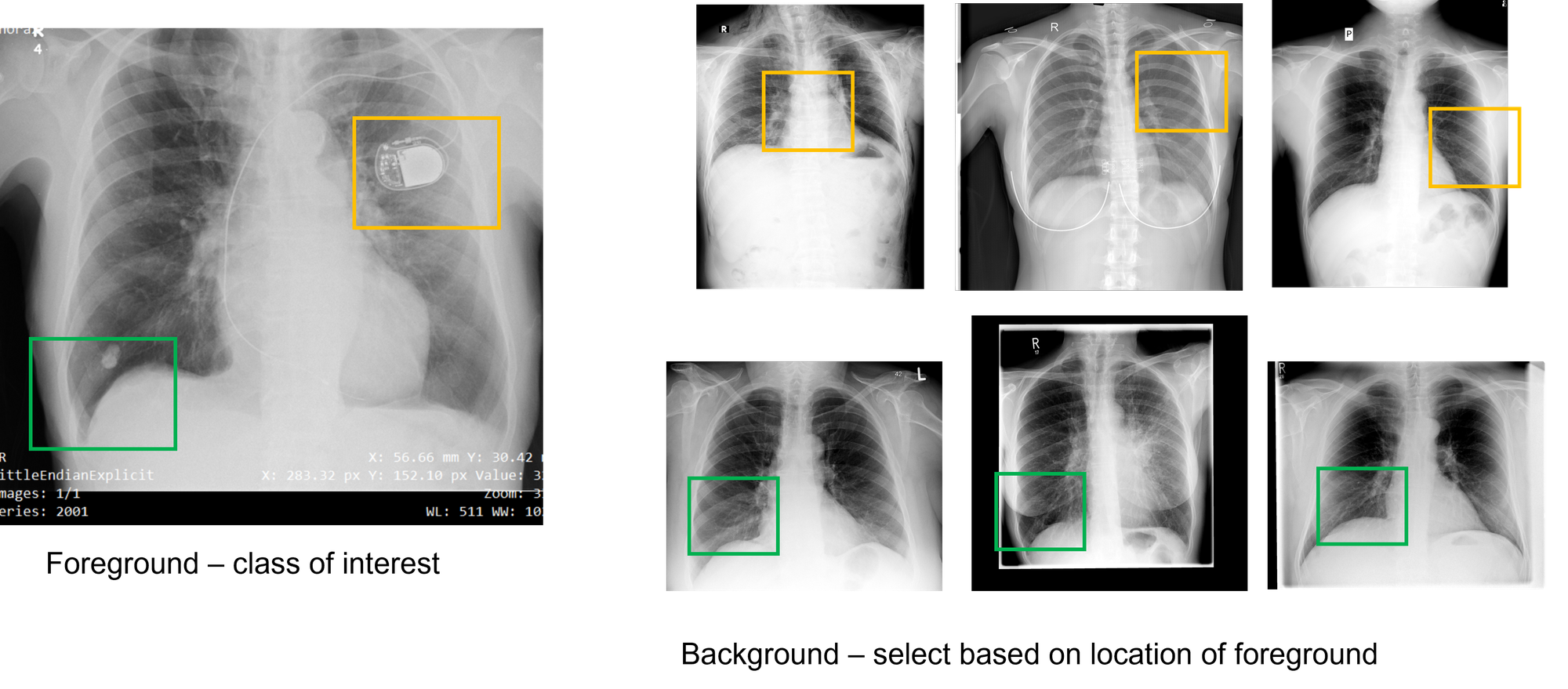}
    \caption{Diagram to show how background areas are selected based on the finding of interest.}
    \label{fig:bkgnd_select}
\end{figure}

Moreover, CXR images have an advantage over natural images, as they are typically well-aligned. Although the location of each finding may vary, the findings tend to appear in specific anatomical regions. Using these characteristics, we select backgrounds based on the prior distribution of findings within our minimal dataset. This approach improves model accuracy by enabling it to distinguish between normal and diseased anatomy. As an example, Fig. \ref{fig:bkgnd_select} highlights the selection of the foreground and background bounding box regions based on the prior location of pleural effusion and a device. 

\begin{figure}[t]
    \centering
    \begin{minipage}{0.48\linewidth}
        \centering
        \includegraphics[width=\linewidth]{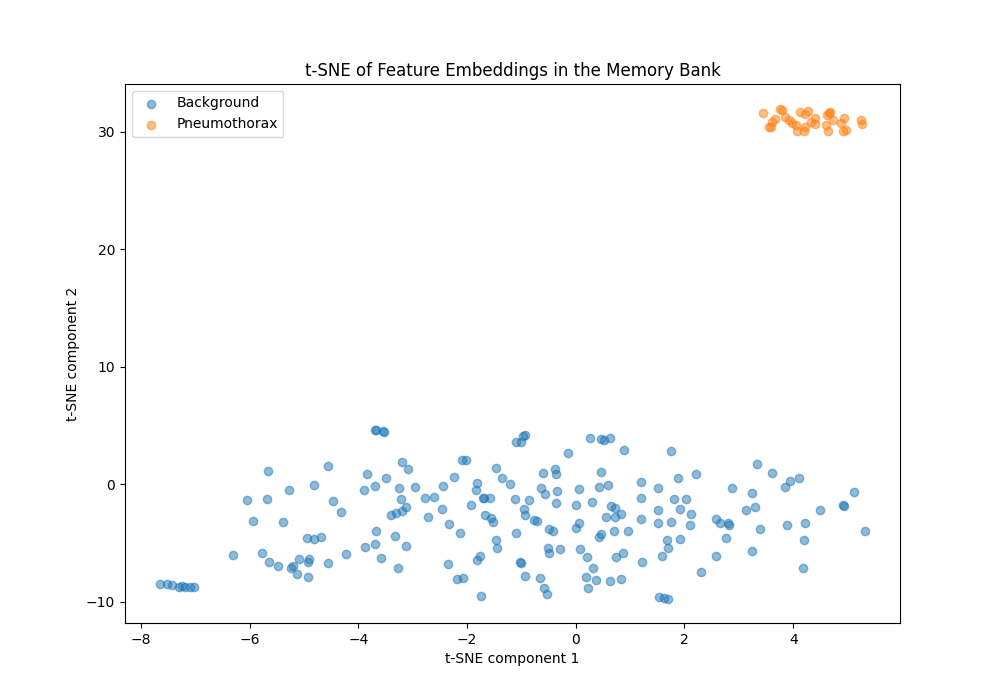}
    \end{minipage}%
    \hspace{0.02\linewidth} 
    \begin{minipage}{0.48\linewidth}
        \centering
        \includegraphics[width=\linewidth]{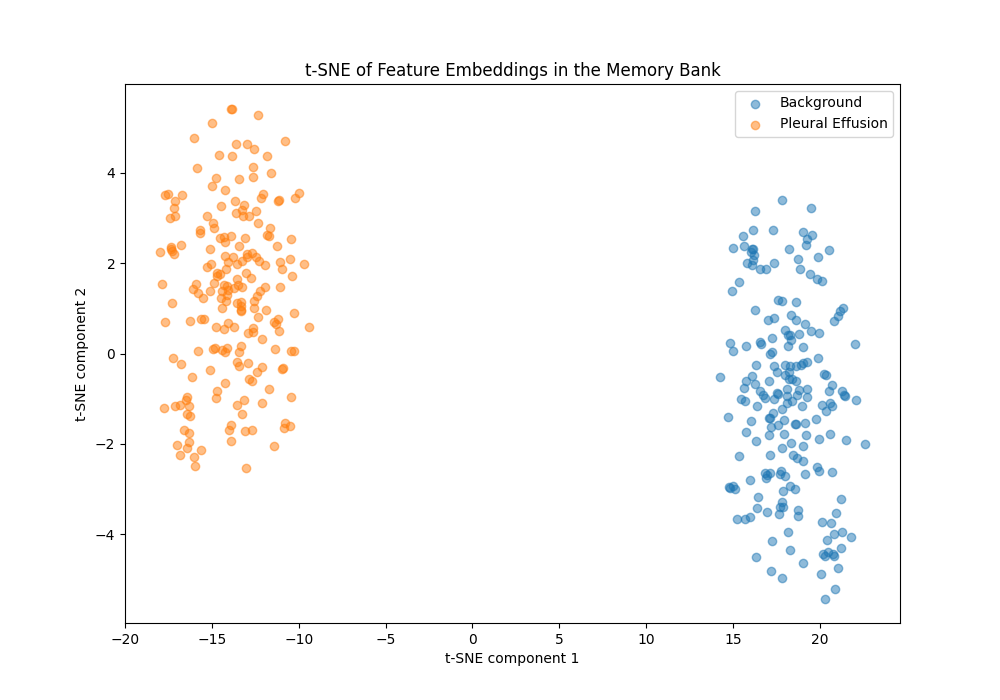}
    \end{minipage}
    \caption{2D t-SNE analysis of pneumothorax (left) and pleural effusion (right) prototypes contrasted with their background samples.}
    \label{fig:tsne_combined}
\end{figure}

Additionally, stage III uses multiple false positive locations as background annotations to further refine the model, thereby enhancing learning and performance metrics. Fig. \ref{fig:tsne_combined} illustrates the effectiveness of exemplar feature extraction and class separation between the foreground-background classes for pneumothorax and pleural effusion through a TSNE \cite{van2008visualizing} analysis of the embeddings stored in the memory bank. 

\section{Experiments}

\subsection{\textbf{Experimental details}}

The experiments were run on a single node with four 40 GB A100 GPUs, using a learning rate of 0.0008 with a linear scheduler under the MMDetection \cite{chen2019mmdetection} framework. An average of 5 runs is recorded. The image and text backbones were frozen. $\mathcal{F}$ is designed as a simple 2-head, 4-layer transformer. The moving average is computed over $L=200$ exemplars. The mean average precision at 50\% IoU ($mAP_{50}$) is reported. 

\paragraph{\textbf{Background selection}}
The image is divided into 4 quadrants, and two background annotations are selected per quadrant, based on the approximate location of the foreground class. We randomly add a 50 pixel top/bottom, left/right perturbation to the selected location along with a 20\% variation in size of the bounding box based on the average size of the foreground class. 
At the final stage III, we further finetune the model with the top 8 misclassified regions (FP) in training images set as background examples. 

\subsection{\textbf{Evaluation metrics}}
We evaluate all models using both classification and localization metrics. For image level classification, the highest probability per image per class is saved to calculate the image level area under the ROC curve (AUC) score. The $mAP_{50}$ metric is used to verify the localization result at higher Intersection-over-Union levels of 50\%. The default $mAP_{50}$ is calculated using the standard coco metric settings provided by the MMDetection \cite{chen2019mmdetection} framework. As a practical metric, we also evaluate the distance-based Free-response Receiver Operating Characteristic (FROC) response of the model \cite{miller1969froc,pham2022accurate,bhat2025patch, 9395510,ghesu2022contrastive,9746079,9091120}. The sensitivity is computed on the basis of the distance between the center of the ground truth and the predicted box \cite{maier2024metrics}. 

\section{Experimental Results}

\subsection{\textbf{Comparison with SOTA} }\label{comparison_results}

Table. \ref{tab:anatomy_detection} indicates the few-shot detection results for the ``heart'' class on the anatomy dataset $A$. The first column presents per-class $mAP_{50}$ for baseline grounding DINO \cite{liu2024grounding}, and the second column for EM-DETR. With just 5 heart-annotated images, EM-DETR boosts the $mAP_{50}$ from 23\% to 92\%. However, a 3-4\% decrease in the $mAP_{50}$ for the middle and lower ribs is observed due to overlapping annotated regions with heart, thus leading to an increase in the cosine similarity and consequent decrease in the classification accuracy. An ablation study in Appendix \ref{fig:heart_ablation} indicates that the $mAP_{50}$ for the heart reaches 99\% with only 15 images. 

\begin{table}[t]
\footnotesize
\centering
\begin{tabular}{lcc}
\toprule
\textbf{Category}     & \textbf{Baseline G.DINO} \cite{liu2024grounding} & \textbf{EM-DETR} \cite{emdetr} \\ 
\midrule
Clavicle           & 0.94              & 0.95                     \\ 
Aorta              & 1.0               & 0.99                     \\  
\textbf{Heart}     & 0.23              & \textbf{0.92}            \\ 
Lungs              & 1.0               & 1.0                      \\ 
Upper ribs         & 1.0               & 1.0                      \\ 
Middle ribs        & 1.0               & 0.97                     \\ 
Lower ribs         & 1.0               & 0.96                     \\ 
Upper spine        & 1.0               & 1.0                      \\ 
Middle spine       & 1.0               & 1.0                      \\ 
Lower spine        & 1.0               & 1.0                      \\ 
\midrule
Average  &  0.82 & \textbf{0.98} \\
\bottomrule
\end{tabular}
\caption{Few shot detection of the heart class with only 5 training images for heart and 80 training images for other classes. The results are depicted in average $mAP_{50}$. The inclusion of the Exemplar Generation module $\mathcal{G}$ results in a decrease in detection performance for classes that overlap with the heart region due to overlapping region boundaries.}
\label{tab:anatomy_detection}
\end{table}

\begin{table}[t]
\footnotesize
\centering
\begin{tabular}{lllcccc}
\toprule
 &{\textbf{Method}} & \textbf{Data \%} & \multicolumn{2}{c}{\textbf{Sensitivity}} & {\textbf{AUC}} & \textbf{$mAP_{50}$}  \\ 
\cmidrule(lr){4-5} 
 && & \textbf{0.25 FP} & \textbf{0.5 FP} && \\ 
 \midrule

\multirow{6}{*}{\rotatebox{90}{\textbf{Ptx}}} & Baseline G.DINO \cite{liu2024grounding}  & 7\% & $0.518\pm\gamma$ & $0.634\pm\gamma$ & $0.837\pm\epsilon$ & $0.258\pm0.05$ \\ 
&RPS \cite{ghesu2022contrastive} & 7\% & 0.518 & 0.642 & 0.877 & -  \\ 
&EM-DETR  \cite{emdetr} & 7\% & \boldmath{$0.885\pm\epsilon$} & \boldmath{$0.932\pm\epsilon$} & \boldmath{$0.957\pm0$} & \boldmath{$0.427\pm0.02$}  \\
\cmidrule(lr){2-7}
&Baseline G.DINO \cite{liu2024grounding} & 100\% & $0.843\pm\gamma$ & $0.925\pm\gamma$ & $0.960\pm\epsilon$  & $0.524\pm0.03$ \\ 
&RPS \cite{ghesu2022contrastive} & 100\% & \textbf{0.970} & \textbf{0.970} & \textbf{0.983} & -  \\                   
&EM-DETR \cite{emdetr} & 100\% & $0.915\pm\epsilon$ & $0.956\pm\epsilon$ & 0.980 & \boldmath{$0.564\pm0.01$}  \\ 
\midrule

\multirow{6}{*}{\rotatebox{90}{\textbf{Pl. effusion}}} & Baseline G.DINO \cite{liu2024grounding} & 7\% & $0.902\pm\gamma$ & $0.960\pm\gamma$ & $0.975\pm\epsilon$ & $0.314\pm0.03$  \\                 
&RPS \cite{ghesu2022contrastive} & 7\% & 0.670 & 0.819 & 0.894 & -   \\
&EM-DETR \cite{emdetr} & 7\% & \boldmath{$0.987\pm\epsilon$} &\boldmath{$0.987\pm\epsilon$} & \boldmath{$0.986\pm\epsilon$} & \boldmath{$0.352\pm0.02$} \\  
\cmidrule(lr){2-7}
&Baseline G.DINO \cite{liu2024grounding} & 100\% &  $0.878\pm\gamma$ & $0.899\pm\gamma$ & $0.989\pm\epsilon$ & $0.389\pm0.02$  \\ 
&RPS \cite{ghesu2022contrastive} & 100\% & \textbf{0.997} & \textbf{0.997} & \textbf{0.993} &  -  \\   
&EM-DETR \cite{emdetr} & 100\% & $0.987\pm\epsilon$ & $0.987\pm\epsilon$ & \boldmath{$0.993\pm\epsilon$} & \boldmath{$0.422\pm0.03$} \\ 

\midrule

\multirow{6}{*}{\rotatebox{90}{\textbf{Lesion\textsuperscript{\textdagger}}}} &Baseline G.DINO \cite{liu2024grounding} & 7\% & $0.309\pm\gamma$ & $0.368\pm\gamma$  & $0.730\pm\epsilon$ & $0.157\pm0.02$  \\                           
&RPS \cite{ghesu2022contrastive} & 7\% & 0.420 & 0.487 & 0.789 & -   \\ 
&EM-DETR \cite{emdetr} & 7\% &  \boldmath{$0.541\pm\epsilon$} & \boldmath{$0.650\pm\epsilon$} & \boldmath{$0.865\pm\epsilon$} & \boldmath{$0.338\pm0.01$}  \\  
\cmidrule(lr){2-7}
&Baseline G.DINO \cite{liu2024grounding} & 100\% & $0.458\pm\gamma$ & $0.678\pm\gamma$ & $0.901\pm\epsilon$ & \boldmath{$0.527\pm0.02$}  \\ 
&RPS \cite{ghesu2022contrastive} & 100\% & \textbf{0.800} & \textbf{0.890} & \textbf{0.950} & - \\ 
&EM-DETR \cite{emdetr} & 100\% & $0.596\pm\epsilon$ & $0.691\pm\epsilon$ & $0.899\pm\epsilon$ & {$0.514\pm0.02$} \\ 

\bottomrule
\end{tabular}
\caption{Comparison of Sensitivity and AUC Scores for Different Findings at Various False Positive Rates and Dataset Sizes. All experiments conducted with a Swin-B image encoder backbone. In the case of Lesion$\dagger$ we evaluate the highest scale of the multi-scale feature from the Swin-B backbone. For readability we denote the standard deviation as $\epsilon \leq 0.01$ and $\gamma \leq 0.1$.}
\label{tab:findings_comparison}
\end{table}

Following this, EM-DETR is tested on a task involving detecting three specific abnormalities or findings. Table. \ref{tab:anatomy_detection} presents the detection results when using only 7\% of the training data for pleural effusion, pneumothorax, and lesions. The sensitivites at 0.25 FP and 0.5 FP are noted from the FROC curves along with the classification AUC scores and localization $mAP_{50}$. An average 10–30\% point sensitivity improvement and 2–13\% point AUC increase is observed for all three findings compared to the baseline Grounding DINO model \cite{liu2024grounding} when using only 7\% of annotations. We also discern 4-16\% point improvement in $mAP_{50}$ with 7\% annotations and a corresponding 3-4\% point improvement from the baseline when 100\% of annotations are used. Our results are further compared with our proprietary Reference Product Solution (RPS), demonstrating comparable performance at maximum annotations and superior performance with minimal annotations in all three cases. 

We observe that EM-DETR achieves near SOTA sensitivity and classification results for minimal-shot pleural effusion detection, unlike pneumothorax and lesions. This is attributed to the closely clustered latent space distribution and the consistent positioning of pleural effusion within CXR images as seen in Fig.~\ref{fig:tsne_combined}. Analysis of the failure cases in Fig.~\ref{fig:failure_cases} shows that pleural effusion is typically and correctly detected at the lung bases even when bounding box overlap is low. 

\begin{figure}[ht]
    \centering
    \begin{subfigure}[b]{0.48\textwidth}
        \centering
        \includegraphics[width=\textwidth]{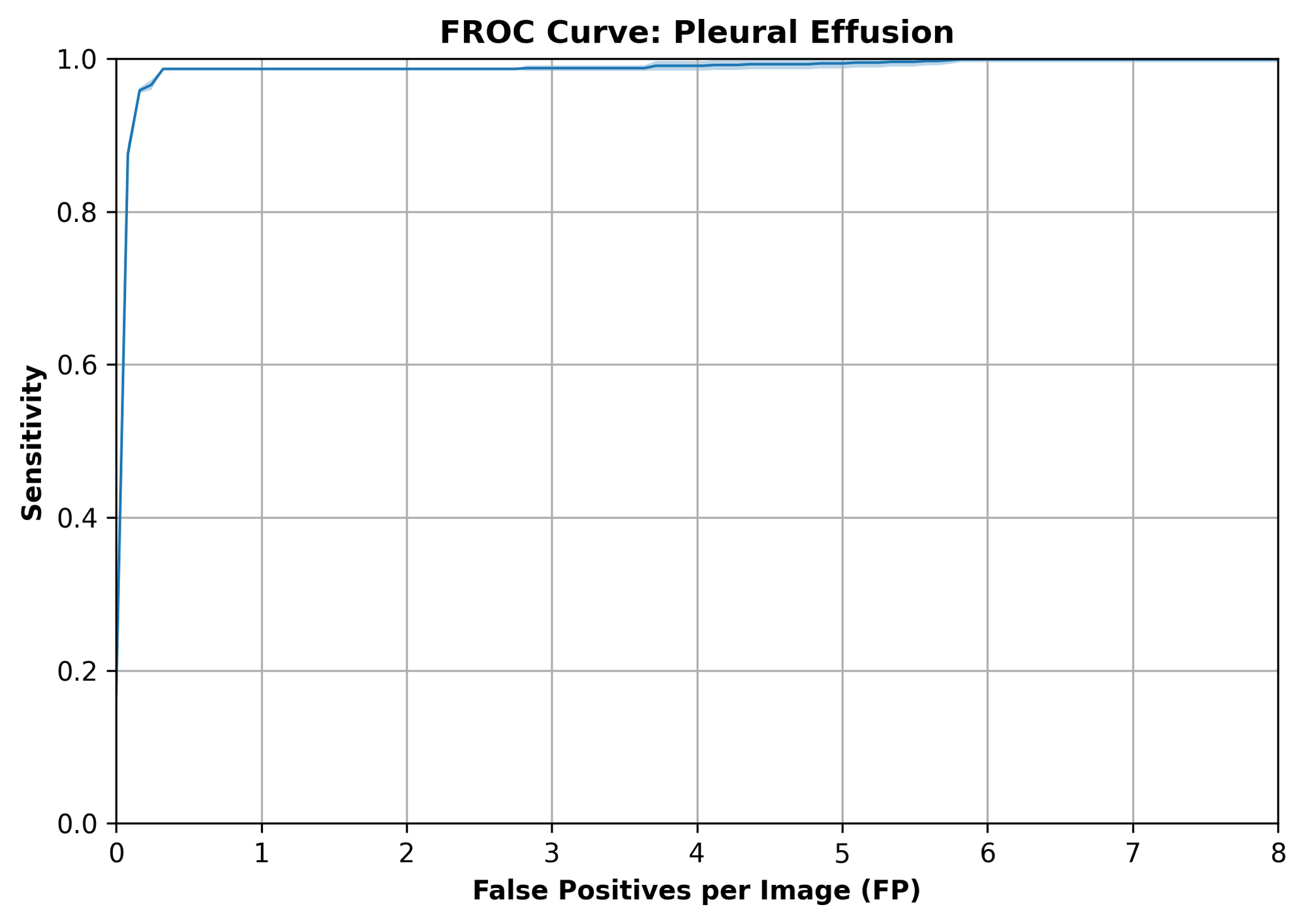}
        \caption{FROC curve for pleural effusion.}
        \label{fig:froc_a}
    \end{subfigure}
    \hfill
    \begin{subfigure}[b]{0.48\textwidth}
        \centering
        \includegraphics[width=\textwidth]{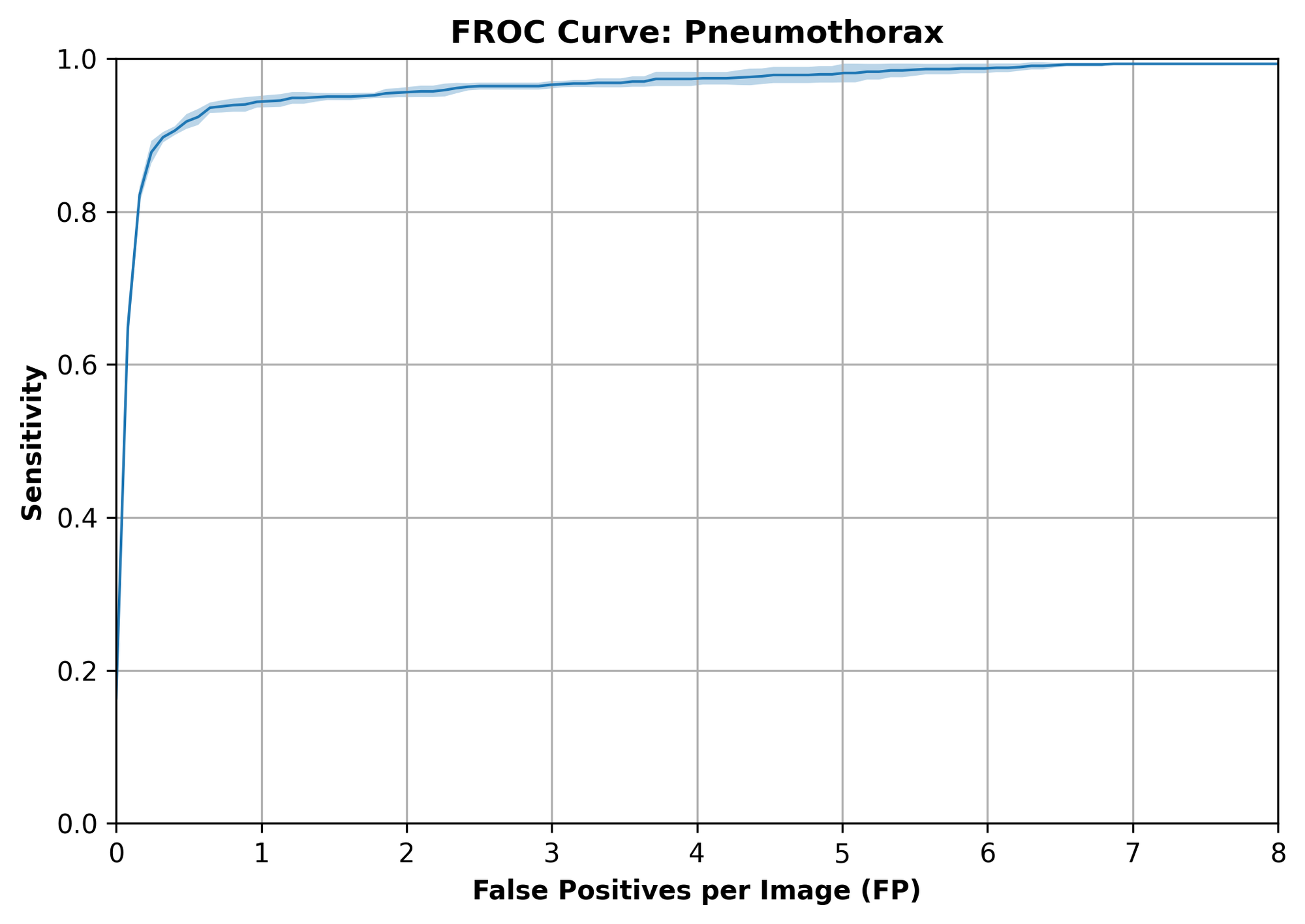}
        \caption{FROC curve for pneumothorax}
        \label{fig:froc_b}
    \end{subfigure}
    \caption{Example FROC curves from EM-DETR for a pair of findings, displaying the average and standard deviation when finetuned on less than 10\% of the training data.}
    \label{fig:froc_comparison}
\end{figure}

Conversely, pneumothorax exhibits much higher variation in appearance and location, which is reflected in the ablation study (Section~\ref{ablation_section}). Increasing the number of pneumothorax annotations from 200 to 400 substantially improves both classification and localization, approaching SOTA performance (Table~\ref{tab:datasize_impact}). This underscores the importance of sample diversity for certain pathologies and suggests that uniform random sampling may be insufficient for findings with heterogeneous presentation. Fig.~\ref{fig:froc_comparison} shows the resulting FROC curves for pleural effusion (left) and pneumothorax (right) under $<$10\% training data.

As an additional experiment for lesion detection, we evaluated the pretrained model using only the highest-resolution scale of the multi-scale features. Despite this simplification, the model still outperformed the SOTA FROC results. Given that lesions are complex abnormalities, careful selection of background samples is necessary, as evidenced by the performance gap with 100\% annotations.



\begin{table}[t]
\footnotesize
\centering
\begin{tabular}{lccc}
\toprule
\textbf{Method (Dataset size)} & \multicolumn{2}{c}{\textbf{Sensitivity}} & \textbf{AUC}\\
 \cmidrule(lr){2-3} 
 & \textbf{0.25FP} & \textbf{0.5FP}  &  \\
\midrule
{EM-DETR (7\%)} \cite{emdetr} & \boldmath{$0.722\pm0.02$}* & \boldmath{$0.833\pm0.01$} & \boldmath{$0.909\pm0.01$}*  \\
{Grounding DINO (7\%)} \cite{liu2024grounding} & $0.556\pm0.06$ & $0.777\pm0.13$ & $0.853\pm0.02$ \\
{RPS (7\%)} \cite{ghesu2022contrastive} & 0.278 & 0.333 & 0.682   \\ 
Deformable DETR (7\%) \cite{carion2020end} & 0.271 & 0.542 & 0.524 \\
\midrule
Deformable DETR (100\%) \cite{carion2020end} &\textbf{0.944} & \textbf{0.944} & \textbf{0.995} \\
{RPS (100\%)} \cite{ghesu2022contrastive} & 0.889 & 0.889 & {0.967}  \\
{EM-DETR (100\%)} \cite{emdetr} & $0.889\pm0.02$* & \boldmath{$0.944\pm0.03$}* & $0.965\pm0.01$* \\ 
{Grounding DINO (100\%)} \cite{liu2024grounding} & $0.833\pm0.04$ & $0.833\pm0.03$ & $0.897\pm0.02$  \\
{RetinaNet (100\%)} \cite{lin2017focal} &- &-& 0.865 \\
Meta-DETR$^\dagger$ (50\%) \cite{zhang2022meta} & - & - & 0.857 \\
{RTMDet (100\%) \cite{lyu2022rtmdet}} &-&-& 0.806 \\
Deformable DETR$^\dagger$ (100\%) \cite{carion2020end} & - & - & 0.729 \\
{Faster RCNN (100\%) \cite{girshick2015fast}} &-&-& 0.496 \\

\bottomrule
\end{tabular} 
\caption{{Comparison of pneumothorax performance with various SOTA models. $\dagger$ indicates within-distribution tests where the model was trained and evaluated on VinDR-CXR dataset. * indicates statistical significance with p $<$ 0.05 when compared to the baseline. Standard deviation provided where available.}}
\label{tab:vindr_zeroshot}
\end{table}

We further evaluate our finetuned EM-DETR model on the VinDR-CXR \cite{nguyen2022vindr} dataset and compare its performance with an in-house RPS solution. The models are finetuned only on the $\mathcal{B_P}$ and $\mathcal{B_F}$ datasets and not further finetuned on VinDR-CXR, allowing us to assess their out-of-distribution (OOD) performance. As shown in Table~\ref{tab:vindr_zeroshot}, EM-DETR outperforms the SOTA baseline when trained with fewer annotations and achieves higher sensitivity at 0.25, 0.5 false positives (FP) per image when using 100\% of the available annotations. These results highlight the efficiency and robustness of EM-DETR in detecting CXR abnormalities even on OOD datasets with minimal supervision. 
For additional context, we also report comparisons with other detection methods. These results serve only as reference points, since our primary evaluation focuses on minimal-shot generalization. Additionally, Deformable DETR is evaluated as a within-distribution and out-of-distribution case. The  model outperforms the RPS when trained on the internal dataset $B$ as it has more pneumothorax training samples (1868) than the VinDR-CXR dataset (96). However, under the minimal-shot configuration, EM-DETR achieves the best performance. Meta-DETR, designed for a much smaller k-shot setting, is trained with only 50\% of VinDR-CXR pneumothorax annotations, as using more samples leads to overfitting and lower performance.

\subsection{\textbf{Ablation Studies}}\label{ablation_section}

In this section, we conduct a few ablation studies to understand the potential, limitations, and impact of the components of EM-DETR.

\begin{table}[t]
\footnotesize
\centering
\begin{tabular}{lcccc}
\toprule
\textbf{Configuration} & \multicolumn{2}{c}{\textbf{Sensitivity}} &  \textbf{AUC}&  \textbf{$mAP_{50}$}\\ \cmidrule(lr){2-3} 
 & \textbf{0.25FP} & \textbf{0.5FP}  & & \\
\midrule
Baseline & $0.518\pm0.01$ & $0.634\pm0.01$ & $0.837\pm0.01$ & $0.258\pm0.11$ \\ 
Baseline+bkgnd($B$) & $0.611\pm0.01$ & $0.756\pm0.01$ & $0.902\pm0.01$ & $0.331\pm0.01$\\
Baseline + B+$\mathcal{G}$ & $0.728\pm0.02$ & $0.806\pm0.02$ & $0.939\pm0.00$ & $0.379\pm0.03$ \\
Baseline + $B$+$\mathcal{G}$+ frozen $F$ & $0.766\pm0.01$& $0.878\pm0.02$& $0.937\pm0.00$ & $0.346\pm0.02$ \\
EM-DETR & \boldmath{$0.885\pm0.01$} & \boldmath{$0.932\pm0.01$} & \boldmath{$0.957\pm0.00$} & \boldmath{$0.427\pm0.02$} \\
\bottomrule
\end{tabular}
\caption{Impact of each module on classification and detection performance for pneumothorax. Experiments are conducted using the Swin-Base backbone within the MMDetection framework, with default data augmentations. The first row shows the baseline performance. The second row adds $B$, indicating inclusion of background annotations in the minimal-shot training file. The third row adds $\mathcal{G}$, denoting the exemplar feature extraction module. In the fourth row, frozen $F$ indicates that the attention pooling module is frozen after convergence to further refine the decoder. The last row is the full EM-DETR pipeline finetuned after pretraining on base classes.}
\label{tab:feature_impact}
\end{table}

Table. \ref{tab:feature_impact} lists the impact of each of the proposed modules, where the first row denotes the baseline grounding DINO \cite{liu2024grounding} performance when the model is trained with 2 or more classes. The second row shows the effect of including background annotations and training the model on a background vs. foreground task. This is followed by the impact of our proposed feature extraction module $\mathcal{G}$. In the next experiment, we freeze the attention pooling network $F$ after convergence, to further refine the decoder for an additional 10 epochs. This experiment is conducted to showcase that training both the decoder and feature extractor simultaneously is recognized for its effect on performance\cite{zhang2022meta}. In this scenario, although $mAP_{50}$ decreases, the FROC sensitivities improve. Freezing the encoder has a similar impact on performance as freezing $F$.In the last experiment, we use the iterative training strategy, where we pretrain with $B_P$ and finetune with $B_F$. (Stage I and Stage II training). 

\begin{table}[t]
\footnotesize
\centering
\begin{tabular}{lcccc}
\toprule
\textbf{Training data} & \multicolumn{2}{c}{\textbf{Sensitivity}} & \textbf{AUC} & \textbf{$mAP_{50}$}\\ \cmidrule(lr){2-3} 
 & \textbf{0.25FP} & \textbf{0.5FP}  & &  \\
\midrule
{50 ptx + bkgnd} & $0.679\pm0.01$ & $0.773\pm0.01$ & 0.914  & $0.262\pm0.24$ \\ 
{50 ptx + FP as bkgnd}  & $0.753\pm0.01$ & $0.856\pm0.01$ & 0.920  & $0.315\pm0.02$ \\ 
{100 ptx + bkgnd}          & $0.823\pm0.01$ & $0.893\pm0.01$ & 0.951  & $0.340\pm0.01$ \\ 
{200 ptx + bkgnd}          & $0.885\pm0.01$ & $0.932\pm0.01$ & 0.957  & $0.427\pm0.02$ \\ 
{400 ptx + bkgnd}          & $0.897\pm0.01$ & $0.934\pm0.00$ & 0.964 & $0.506\pm0.02$ \\ 
{100\% ptx, no bkgnd}    & 0.$904\pm0.01$ & $0.925\pm0.01$ & 0.977  & $0.512\pm0.03$ \\ 
{100\% ptx + bkgnd} & \boldmath{$0.915\pm0.01$} & \boldmath{$0.956\pm0.01$} & \boldmath{$0.980$} & \boldmath{$0.564\pm0.01$} \\ 
\bottomrule
\end{tabular}
\caption{Comparison of pneumothorax performance with different number of positive annotations. The negative annotations are kept equal to the positive sample set in each experiment. The second row shows the impact of further retraining the model with False Positives as background annotations in a second stage of finetuning.}
\label{tab:datasize_impact}
\end{table}

Table. \ref{tab:datasize_impact} demonstrates the impact of different number of pneumothorax annotations. The results show a trend of improvement with more annotations, achieving near SOTA RPS levels with just 14\% of annotations. Additionally in the case of 50 positive annotations, we improve results with the Stage III training phase as described in Section~\ref{iterative_subsection}. This refinemenet yields a 7-8\% point improvement in sensitivity and an improvement of 5.3\% point in $mAP_{50}$. The final two rows illustrate the improved performance of training EM-DETR with and without background samples, underscoring the significance of high-quality contrasting samples for effective learning.

\subsection{\textbf{Qualitative results}}

\begin{figure}[ht]
    \centering
    \includegraphics[width=1\linewidth]{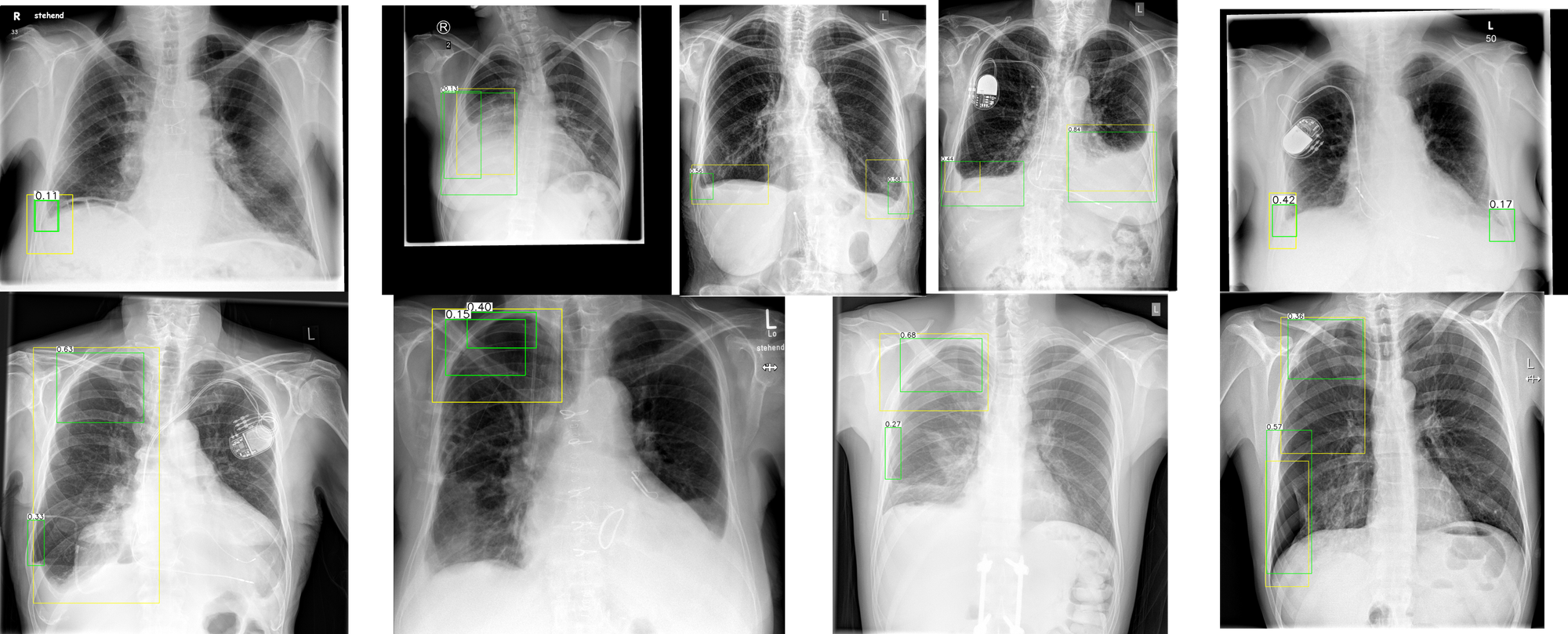}
    \caption{Top row illustrates examples of top 2 predictions for pleural effusion while bottom row illustrates examples of top 2 predictions for pneumothorax. The groundtruth is in yellow and predictions in green. We observe detected regions are more tightly localized than ground truth, which may span multiple nearby findings of same class, leading to an artificial reduction in mAP.}
    \label{fig:qual_examples}
\end{figure}

Fig.~\ref{fig:qual_examples} highlights a few examples of pleural effusion (top row) and pneumothorax detections (bottom row). In some cases, the detected regions are more tightly localized than the ground truth, which often encompasses broader areas or multiple adjacent instances of findings. This mismatch leads to an underestimation of performance in quantitative metrics such as mAP. Since DETR generates multiple hypothesis regions that overlap the target ROI and uses contrastive cosine similarity to match, several overlapping boxes may be valid positives for the same finding. However, these additional detections are counted as false positives by the evaluation framework, thereby reducing the mAP score as well.

\begin{figure}
    \centering
    \includegraphics[width=1\linewidth]{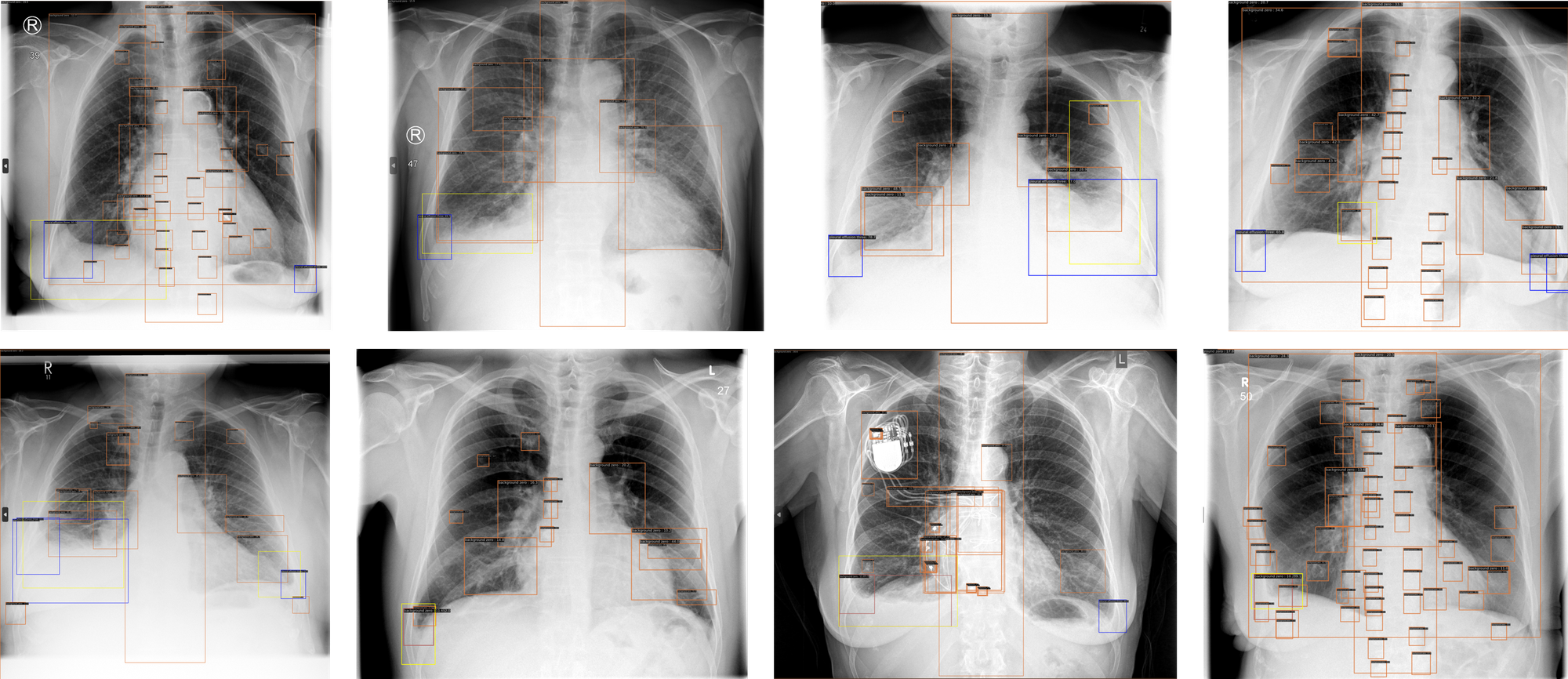}
    \caption{Example images of incorrect detections of pleural effusion on the test dataset. Yellow boxes mark the groundtruth annotation, while blue indicates the predicted pleural effusion with the probability score. The background regions are marked in orange and reflect the training distribution of background annotations.}
    \label{fig:failure_cases}
\end{figure}

Fig.~\ref{fig:failure_cases} illustrates failure cases for pleural effusion detection. In most examples, the model correctly localizes the costophrenic angles and attributes the findings to the appropriate lung. However, a few images show instances where true pleural effusion regions are misclassified as background. Despite these occasional errors, the overall high AUC score suggests that the model accurately identifies abnormal regions in the majority of cases.

\section{Discussion and Conclusion}

EM-DETR achieves near state-of-the-art results on chest radiographs using fewer than 10\% of available annotations, demonstrating a scalable approach for handling long-tailed class distributions with limited annotations. By leveraging few-shot learning to create exemplar or prototype embeddings, our model effectively differentiates between medical findings and anatomical structures, addressing challenges unique to medical imaging. Moreover, the framework is structured to accommodate the addition of new disease classes with limited retraining, laying the groundwork for future studies in continuous learning. The method also incorporates domain-aware negative sampling, which empirically improves detection of rare classes in our few-shot setting. Our previous conference work demonstrated the generalizability of EM-DETR across multiple imaging domains, including mammography and stenosis detection. In this submission, we focus specifically on the minimal-annotation aspect and its effectiveness for chest radiographs. Evaluation across multiple thoracic findings on both proprietary and public datasets demonstrates the robustness and practical utility of the approach.

While EM-DETR demonstrates strong performance under minimal-anno\-tation settings, there are a few limitations to discuss. The current implementation of EM-DETR is memory-intensive, as it maintains a separate list of patch embeddings and interleaves text and feature embeddings in a naive manner. Moreover, the text encoder functions primarily as a pseudo class identifier to leverage Grounding DINO pretraining, rather than contributing meaningful linguistic grounding. Future work will explore more memory-efficient integration strategies, and also extending text tokens with layer-specific feature embeddings from the multi-scale Swin backbone, enabling finer-grained feature matching and reducing ambiguity between related findings. We plan to investigate the implementation of various background prototypes to further distinguish anatomical priors. The method is also sensitive to annotation noise, as incomplete labeling of co-occurring findings can negatively affect the contrastive loss and lead to performance degradation. Addressing these aspects through improved annotation completeness and architectural refinement will be key to further enhancing EM-DETR’s scalability and robustness.

Together, these insights underscore both the promise and current boundaries of EM-DETR in practical deployment. In conclusion, despite the current limitations, which are primarily related to implementation and memory efficiency, extensive pretraining offers the potential to further boost detection capabilities, positioning EM-DETR as a promising step toward scalable and efficient diagnostic methods in healthcare.


\section{Declaration}

During the preparation of this work the author(s) used AI tools in order to improve grammar and readability. After using this tool/service, the author(s) reviewed and edited the content as needed and take(s) full responsibility for the content of the published article.

\section{Acknowledgment}

This study is funded by Siemens-Healthineers.

\appendix
\section{}
\label{app_1}

\begin{figure}[ht]
    \centering
    \includegraphics[width=0.8\linewidth]{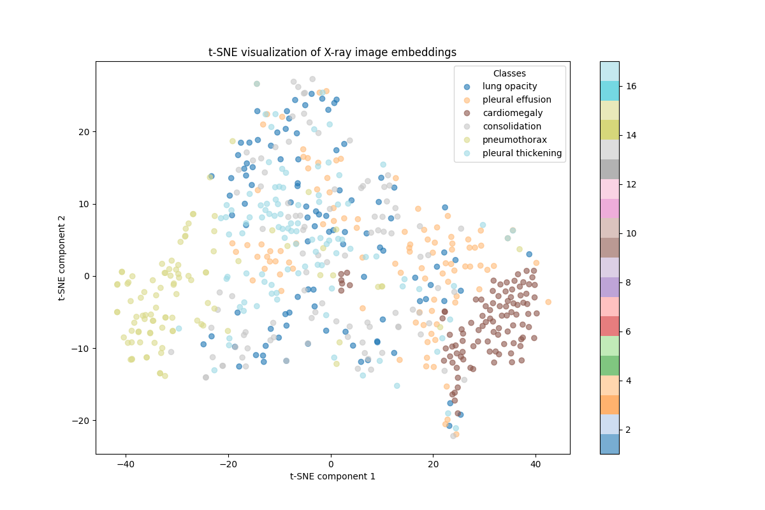}
    \caption{2D TSNE analysis of feature embeddings per class when input to a torchxray vision backbone}
    \label{fig:classifier_tsne}
\end{figure}

Fig. \ref{fig:classifier_tsne} illustrates the 2D TSNE \cite{van2008visualizing} analysis of image backbone embeddings for the regions of interest across each class, when processed using the robust TorchXRayVision model\cite{Cohen2022xrv}.

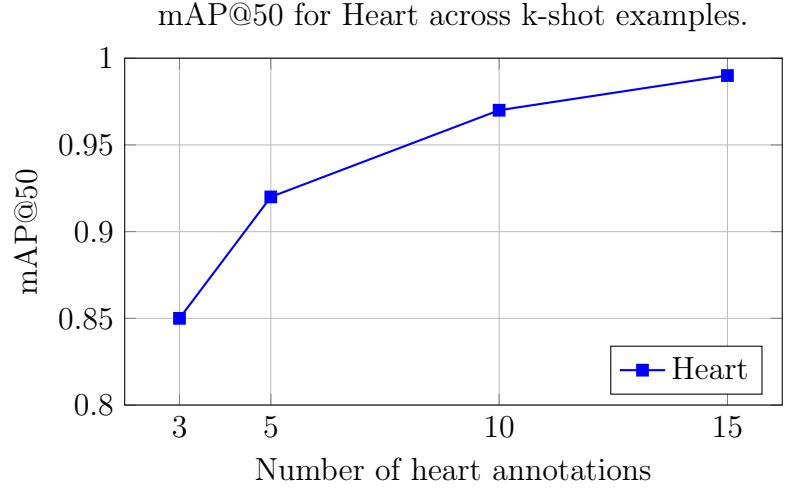
\begin{figure}[h!]
\centering
\begin{tikzpicture}
\begin{axis}[
    xlabel={Number of heart annotations},
    ylabel={mAP@50},
    ymin=0.8, ymax=1.0,
    xtick={3,5,10,15},
    ytick={0.80, 0.85, 0.90, 0.95, 1.00},
    legend pos=south east,
    grid=both,
    width=0.75\textwidth,
    height=0.45\textwidth,
    title={mAP@50 for Heart across k-shot examples.}
]
\addplot[
    color=blue,
    mark=square*,
    thick
] coordinates {
    (3, 0.85)
    (5, 0.92)
    (10, 0.97)
    (15, 0.99)
};
\addlegendentry{Heart}
\end{axis}
\end{tikzpicture}
\caption{mAP@50 performance for the Heart class across different few-shot settings.}\label{fig:heart_ablation}
\end{figure}

Fig.~\ref{fig:heart_ablation} presents an ablation study with varying number of few-shot references images for the heart in the anatomy dataset. 

\newpage
\bibliographystyle{elsarticle-num} 
\bibliography{media}
\end{document}